\documentclass[letterpaper,journal]{IEEEtran}
\usepackage{amsmath,amsfonts}
\usepackage{bm}
\usepackage{algorithmic}
\usepackage{algorithm}
\usepackage{array}
\usepackage{booktabs}
\usepackage[caption=false,font=normalsize,labelfont=sf,textfont=sf]{subfig}
\usepackage{textcomp}
\usepackage{stfloats}
\usepackage{url}
\usepackage{verbatim}
\usepackage{graphicx}
\usepackage{cite}
\usepackage{color}
\usepackage{xcolor}
\usepackage{colortbl}
\usepackage{makecell}
\usepackage{bbding}
\usepackage{amssymb}
\usepackage{threeparttable} 
\usepackage{hyperref}
\hypersetup{
	colorlinks=true,
	linkcolor=blue,
	filecolor=green,      
	urlcolor=red,
	citecolor=cyan,
}

\newcommand{\first}[1]{%
    \cellcolor{green!30}%
    \bfseries #1%
}

\newcommand{\second}[1]{%
    \cellcolor{green!15}%
    \underline{#1}%
}

\newcommand{\third}[1]{%
    \cellcolor{yellow!15}%
    {#1}%
}

\begin{document}

\title{ROEVO: Robust Organized Edge Feature-based Visual Odometry Using RGB-D Cameras}

\author{Mingrui Liu$^{1}$, Xingxing Zuo$^{2}$, Renlang Huang$^{1}$, Minglei Zhao$^{1}$, Jiming Chen$^{1}$, and Liang Li$^{1}$

\thanks{\textit{Corresponding authors: Liang Li and Jiming Chen.}}
\thanks{$^{1}$ College of Control Science and Engineering, Zhejiang University, Hangzhou, 310027, China.}
\thanks{$^{2}$ Department of Robotics, Mohamed Bin Zayed University of Artificial Intelligence (MBZUAI), Abu Dhabi, PO Box 7909, UAE.}
}
        
\maketitle

\begin{abstract}
This work presents a visual odometry (VO) system that leverages image edge features. Edges are spatially expressive cues commonly present across diverse environments, offering rich textural and structural information. However, existing edge-based VO methods often fail to fully exploit this potential.  To this end, we introduce a novel feature representation termed \textit{organized edges}, which transforms disjoint edge pixels into sequentialized clusters, enabling more effective retention and utilization of the underlying textural and structural information.
Another nice property of this formulation is that organized edges can perform edge-level association across multiple frames, enabling the establishment of a co-visibility graph. To achieve precise and efficient pose estimation, we propose a range of particularly designed tracking and joint optimization methods based on the characteristics of organized edges. For tracking, we formulate edge-wise rather than pixel-wise residuals to achieve robust and accurate inter-frame registration. For joint optimization, we introduce a novel shape-preserving edge-fitting method and an organized edge-based Bundle Adjustment (BA) approach, which decomposes the traditional BA problem into fitting and registration to preserve the structural integrity.
Based on these novel techniques, we develop a complete VO system that exclusively employs organized edge features, achieving efficient tracking and precise local mapping. Extensive experiments demonstrate its accuracy and robustness in indoor environments, outperforming or achieving comparable performance to state-of-the-art methods. The source code is publicly available at {\ttfamily https://github.com/liumingrui814/ROEVO}.
\end{abstract}


\section{Introduction}\label{intro}

\IEEEPARstart{P}{ose} estimation using vision sensors has been an active field of research over the past two decades. In recent years, several state-of-the-art Visual Odometry (VO) \cite{dso, svo} and Visual Simultaneous Localization and Mapping (VSLAM) algorithms\cite{orbslam1,lsdslam, monoslam, ptam} have emerged. These algorithms strive to achieve high accuracy and robustness of pose estimation across diverse environments while ensuring real-time performance on general computing devices such as personal computers or robotic processors. The ability to provide high-quality pose esfine time:18.6219
timates enables vision-based sensors to play a crucial role in various robotic and computer vision applications. 

State-of-the-art VO/VSLAM systems can be further categorized into three main groups based on differences in feature extraction, association, and optimization: feature-based methods\cite{orbslam1,orbslam2}, direct methods\cite{dso, lsdslam}, and semi-direct methods\cite{svo}. Feature-based methods rely on repeatable point descriptors\cite{brief, orb} for data association and state estimation. These methods can establish reliable correspondences even with wide baselines, achieving stable and accurate pose estimation in environments with rich textures. Algorithms in \cite{plslam, plsvo, cube_slam, QuadricSLAM} extend it by incorporating more diverse sparse features like lines\cite{lsd,lbd} or semantic objects to assist point features, enabling pose estimation to utilize more concrete features beyond pixels. However, due to fluctuations in image streams and the sparsity of observations, feature-based methods suffer from significant observation noise. This limitation makes it difficult for them to achieve good estimations in texture-poor environments. Although these methods mitigate the impact of observational noise on the estimation results through an intricately designed back-end employing bundle adjustment (BA)\cite{ba}, the performance of BA is also easily affected by the lack of robust co-visibility relations in texture-sparse environments.

Direct methods \cite{dvoslam, dso,lsdslam} bypass the pixel-wise feature association and instead directly optimize the photometric errors for pose estimation. These approaches perform better in texture-less environments than feature-based methods. However, due to the lack of explicit data associations and limited convergence domains, establishing associations over large baselines is challenging for direct methods, which affects their widespread usage. Semi-direct methods \cite{svo} use optical flow\cite{lkoptic} for pixel-level associations, effectively balancing efficiency and association quality by leveraging both intensity and texture information in images. However, tracking points on lines or edges using optical flow can lead to one-dimensional ambiguity along these structures \cite{EDPLVO},  which makes it difficult for traditional optical flow methods to fully leverage these structural details.

In this paper, we focus on an edge-based VO system, aiming to leverage edges in the image as a type of more representative feature to combine intensity, structural, and textural information from images. By integrating these aspects, we aim to develop a more robust and accurate VO system. Edges with various shapes are commonly observed in indoor environments and naturally carry rich structural and textural information. They exhibit stronger spatial representations compared to point features. Moreover, edges are denser and have a lower signal-to-noise ratio than sparse points, making them suitable for estimating camera motion and potentially building lightweight maps.

Despite the well-established methods for extracting edge pixels from images and efforts in previous works to utilize edge information for VO systems \cite{cannyvo,re_slam, reEvo, edge_slam}, these approaches often fail to organize the edge pixels into a more elegant form effectively. Instead, edges are typically treated as scattered, irrelevant pixels.  This results in the loss of substantial structural and textural information contained within the edges. Furthermore, the loose arrangement of pixels makes feature association and joint optimization across multiple images difficult. These limitations contribute to the underperformance of edge-based methods and hinder their practicability.

To address these issues, we propose a method for effectively organizing edges to preserve diverse structural and textural information within them. The edge features extracted in this way are referred to as \textit{organized edges}, which enhance the performance of edge features in both association and optimization. Building upon this concept, we have developed a robust and precise VO system that utilizes organized edges as the only feature. This framework presents a novel perspective on utilizing edge features within VO/VSLAM systems.

The contributions of this paper are outlined as follows:
\begin{itemize}
\item{We develop a robust and precise RGB-D visual odometry system that exclusively utilizes organized edges as features, with tracking and local mapping modules specifically designed.}
\item{We present a novel image edge representation, termed \textit{organized edges}, which enables the full exploitation of the texture and structural information embedded in edge features while supporting incremental edge stitching. We also propose an efficient method to extract organized edges expediently.}
\item{We design a novel method for associating and merging these organized edges across multiple frames, with a further joint optimization approach for organized edges within a fused local map.}
\item{We evaluate our algorithm on a wide range of public datasets and custom-built scenes, demonstrating its accuracy in general scenarios and robustness in sparse-texture environments.}
\end{itemize}

\section{Related Work}
In this section, we primarily review the utilization of free-form image edge information in VO/VSLAM systems. Existing methods can be categorized into two main approaches based on how edge information is used: methods that rely on spatial curves and those that directly utilize edge pixels.

\subsection{Spatial Curves-based Methods}

Utilizing spatial curves for structure-from-motion is an important branch of geometric computer vision, with substantial early works in this area. These studies parameterize a 3D spatial curve in some general mathematical forms or as some piecewise function and employ optimization algorithms for curve recovery and camera pose estimation. For instance, Kaminski \textit{et al.} \cite{algecurve} utilized algebraic curves to recover the epipolar geometry, while Kahl \textit{et al.} \cite{varicurve} modeled free-form 3D curves as continuous functions in space and applied calculus of variations for curve reconstruction from registered 2D images. However, these early methods faced limitations due to their mathematical constraints. Specifically, the approach in \cite{algecurve} is sensitive to noise, while the approach in \cite{varicurve} would fail if curves are only partially observed. Other methods describe curves as piecewise functions determined by several control points, such as B-splines \cite{bspline} or Bezier curves \cite{bezier}. However, these approaches tend to prioritize curve recovery and representation with known poses, with less focus on their contribution to pose optimization. A further limitation of these methods is that, although they provide suitable mathematical representations for spatial curves, they do not offer a general method for extracting curves from real-world images that meet these requirements.

Extracting spatial curves with edges generated by image gradients is a promising idea, as edge extraction from images is a well-studied task. However, spatial curves are ordered structures, their sequential information cannot be directly obtained through common edge extraction methods. Existing edge extraction techniques, whether handcrafted edge detectors like \cite{canny,xdog} that utilize the gradient convolution kernel, or Convolutional Neural Network (CNN)-based approaches like\cite{deep_edge}, rely on image convolution. As a result, their final output is typically a binary map composed of edge pixels and non-edge pixels. These output pixels lack sequential or hierarchical relationships and are not organized into well-defined curve features, neglecting the ordered nature of spatial curves, thus making it difficult to utilize curve features in visual odometry used in practice.

Due to the limitations of spatial curves in practical applications, many works have attempted to extract more intuitive and lightweight structures to enhance usability, such as line features \cite{lsd,lbd} or Manhattan world representations \cite{manhattanstereo}. These structures are generally easier to extract from images compared to free-form spatial curves, and their optimization problems can be represented in an elegant form, for instance, straight lines can be easily triangulated with the help of the Plücker coordinates \cite{plk}, while a Manhattan world can be represented by three mutually orthogonal spatial lines \cite{manhattanstereo}, enabling the camera to estimate its rotation based on the world's spatial constraints. VO/VSLAM systems based on these approaches have demonstrated practicality and robustness \cite{structslam, plslam, manhattanslam}. However, these methods restrict the types of features that can be utilized, limiting their effectiveness to specific scenarios.

\subsection{Edge Pixel-based Methods}

In contrast to representing edges as spatial curves, some methods directly treat extracted edge pixels as a set of semi-dense or sparse points, enabling pose estimation using generic methods like registration-based or direct approaches. For example, the method in \cite{edge_slam} utilizes Lucas-Kanade (LK) \cite{lkoptic} optical flow to track edge pixel points and perform tracking and mapping based on the associations. However, using sparse optical flow methods to track edge pixels in texture-deprived environments often encounters dimensionality reduction issues, making it hard to establish pixel-level correspondences. Due to the significant image gradients associated with edge pixels, many direct methods \cite{dso,lsdslam} prioritize selecting pixels along edges to formulate optimization problems. To mitigate the impact of the nonconvexity of the underlying direct objective, several works utilize Euclidean Distance Fields (EDF) \cite{edf} for pose optimization. EDF is a two-dimensional energy function constructed from the foreground pixels of the edge map, where each function value represents the Euclidean distance from the current position to the nearest foreground pixel. By minimizing the energy function derived from reprojected pixels, the transform between consecutive images can be estimated. Since EDF is a differentiable function, it exhibits more favorable properties than photometric error during the optimization. Building on this concept, the method in \cite{edgedirect} uses both photometric error and EDF for pose estimation, while the approach in \cite{revo} directly substitutes the energy function derived from EDF for photometric error. Additionally, the method in \cite{re_slam} incorporates local mapping and loop closure to form a complete SLAM system. However, it is difficult to conduct optimization due to the residual provided by EDF being semi-positive definite. To address this, the method in \cite{cannyvo} modifies EDF to Approximate Nearest Neighbor Field (ANNF), providing vectorized residuals instead of scalar residuals from EDF and achieving better results. Moreover, it proposes a simple association based on the direction of edge gradients during the optimization. The method in \cite{edgevo} further incorporates a feature selection strategy\cite{good_feature} on top of using distance fields, providing a mechanism for data filtering in semi-dense edge point sets.

\begin{figure*}[htbp!]
\centering
\includegraphics[width=0.85\linewidth]{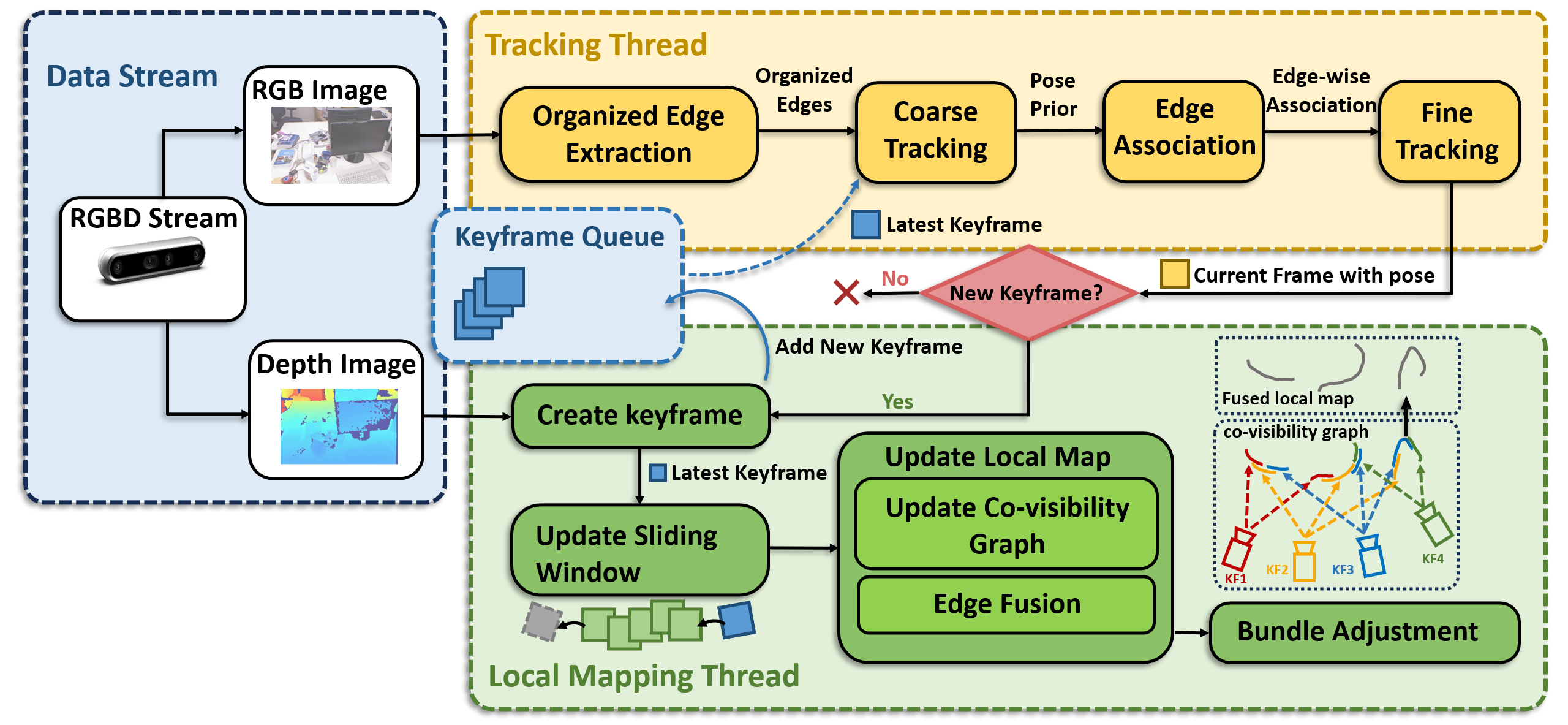}
\caption{Pipeline of the ROEVO system, including the tracking thread marked in yellow and the local mapping thread marked in green. In the tracking thread, organized edges are extracted and pose estimation is performed for each frame, and in the local mapping thread, a local map within a window is maintained and a joint optimization across multiple keyframes is performed.}
\label{pipeline}
\end{figure*}

Despite the numerous advantages of the distance field, constructing a dense distance field still incurs significant computational overhead even with region-growth methods like \cite{edtc}. Furthermore, this dense function involves many unnecessary computations since only a small portion of a real-world image belongs to foreground edges. To address these issues, the method in \cite{iros2018} associates edge points using image gradients and utilizes a KD-tree carrying gradient direction information for more efficient association. On the other hand, the method in \cite{iros2020} clusters edge pixels with consistent pixel gradients and categorizes them based on the orientation of edge clusters for association. Although this approach attempts to estimate trajectories by directly associating clustered edge pixels that contain consistent textual information, the rudimentary clustering methods yield fragmented edge segments. As a result, the utilization of edge information remains loose and suboptimal.

From the aforementioned works, it is evident that while methods based on spatial curves can effectively leverage structural information from edges, they face challenges in edge extraction and are constrained by their mathematical limitations. On the other hand, methods that directly use edge pixels are more practical and adaptable for various shapes of edges. However, the loose representation of edge pixels disrupts the structural information contained within complete edges, making it less conducive for further association and joint optimization. The key to addressing these issues lies in a more organized and robust representation of image edge information. An effective edge representation should preserve the integrity of edges while incorporating their spatial structural information. Only through this approach can the rich structural and textural information embedded in edge features be fully utilized, paving the way for precise and robust edge-based association and optimization.

\section{System Overview}

Our complete RGB-D VO pipeline is illustrated in Fig. \ref{pipeline}. The entire odometry system contains two parallel threads: the tracking thread and the local mapping thread. In the tracking thread, pose estimation is performed for each RGB-D image frame based on the extracted organized edge features described in Section \ref{section_oedge}. 
 Frame-to-frame registration is conducted between the latest keyframe and the current frame to estimate the pose of the current frame. To enhance algorithm efficiency, only the RGB image of the current frame is utilized for extracting organized edges, while keyframes incorporate 3D information provided by depth images. This design formulates the registration problem between keyframes and reference frames as a 3D-2D registration problem.
To balance efficiency and accuracy, we employ a coarse-to-fine strategy, enabling better association of organized edge features while estimating the pose of the current frame. Details of this process are discussed in Section \ref{section_tracking}. In the local mapping thread, we maintain a sliding window of keyframes, and a local map is constructed based on co-visibility relationships among multiple keyframes. Based on these co-visibility relationships, local bundle adjustment (BA) is performed for joint optimization of keyframes within the sliding window. Given that organized edges represent a novel feature type, their association, local map construction, and BA approach all exhibit new characteristics. These aspects are discussed in detail in Section \ref{section_joint}.

\section{Organized Edge Feature}
\label{section_oedge}
In this section, we first illustrate the fundamental concept of organized edge features and how they can carry texture and structural information. Following this, we present a novel method to efficiently extract organized edges.

\subsection{Formulation of Organized Edges}
\begin{figure}[!t]
\centering
\includegraphics[width=0.85\linewidth]{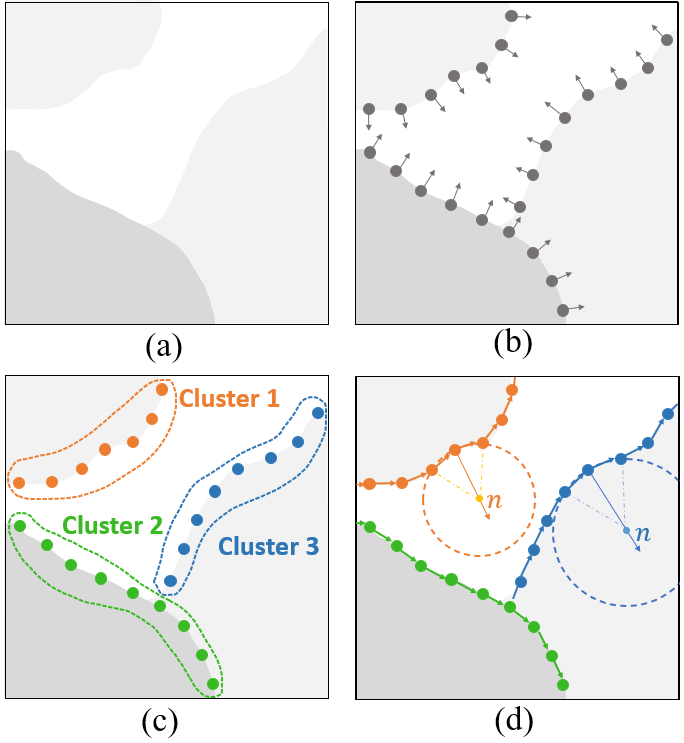}
\caption{The illustration of organized edge features. (a) The raw image;  (b) Raw edges extracted from edge detectors like~\cite{canny}, with arrows representing the image gradients; (c) Clustered edge points based on image gradients; (d) Organized edges after sequentialization, where each pixel's normal vector $n$ can be computed by utilizing its neighboring two edge points.}
\label{idea_odege}
\end{figure}

Given an image as shown in Fig. \ref{idea_odege} (a), generic convolution-based methods can extract a 2D edge mask, as shown in Fig. \ref{idea_odege} (b). At this stage, though edge pixels are extracted, they are just a bunch of loose scattered pixels lacking sequential or hierarchical relationships. To further distinguish these pixels, their texture information needs to be analyzed. For edge pixels, the most direct texture information comes from the pixel patches on either side of the edge as well as the magnitude and direction of image gradients. This implies that edge pixels can be clustered based on their gradient information, as illustrated in Fig. \ref{idea_odege} (c). After clustering, cohesive edges are obtained instead of scattered edge pixels. To further encode the structural context of edge features, we need to calculate the curvatures along these edges. For a curve represented by discrete points, the calculation of curvatures requires knowledge of the sequential order. Given unordered edge clusters depicted in Fig. \ref{idea_odege} (c), a further operation is required to sequentialize these clustered pixels based on their spatial relationships, ultimately resulting in ordered points shown in Fig. \ref{idea_odege} (d). Once the points are ordered whether in 2D or 3D space, the curvature at a given point can be determined using its preceding and succeeding points, enabling a more detailed description of the spatial structure for the edge. 

Through the aforementioned process, by defining edges as ordered sequences of points, textural information is incorporated into the edge cluster, while structural information is embedded in the sequential order of the pixels. In summary, clustering and sequentialization are two pivotal operations for transforming edge features into the desired organized format, enabling the subsequent association and optimization processes to effectively utilize both textural and geometric context embedded within the edges.

\subsection{Organized Edge Extraction}

\begin{figure}[t]
\centering
\includegraphics[width=0.85\linewidth]{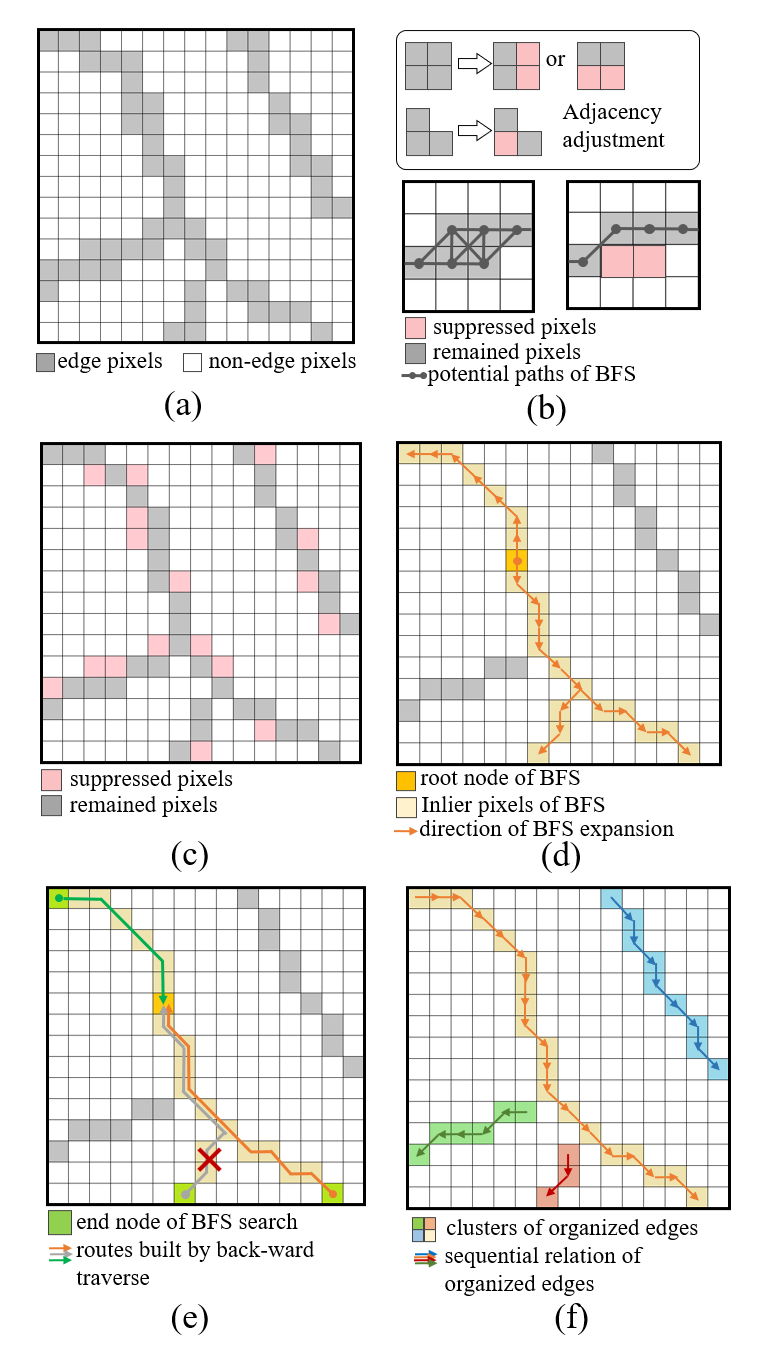}
\caption{The pipeline of our organized edge extraction method. (a) Edges extracted from the Canny algorithm~\cite{canny} (b) Adjacency adjustments on the two pre-defined local pixel patterns, where the pixels marked in pink will be excluded. After suppressing these pixels, BFS will obtain a unique and concise expansion route. (c) Appearance of the edge mask after adjacency adjustment, where the pixels marked in pink are excluded. (d) An edge cluster with expansion routes after oriented BFS.  (e) Paths backtrack from end nodes of the cluster, the gray paths are excluded as they're relatively short, and only the green and orange paths are retained to form the longest ordered sequence of the edge cluster together. (f) The final extracted organized edges.}
\label{extraction_process}
\end{figure}

\begin{figure}[h]
\centering
\includegraphics[width=0.9\linewidth]{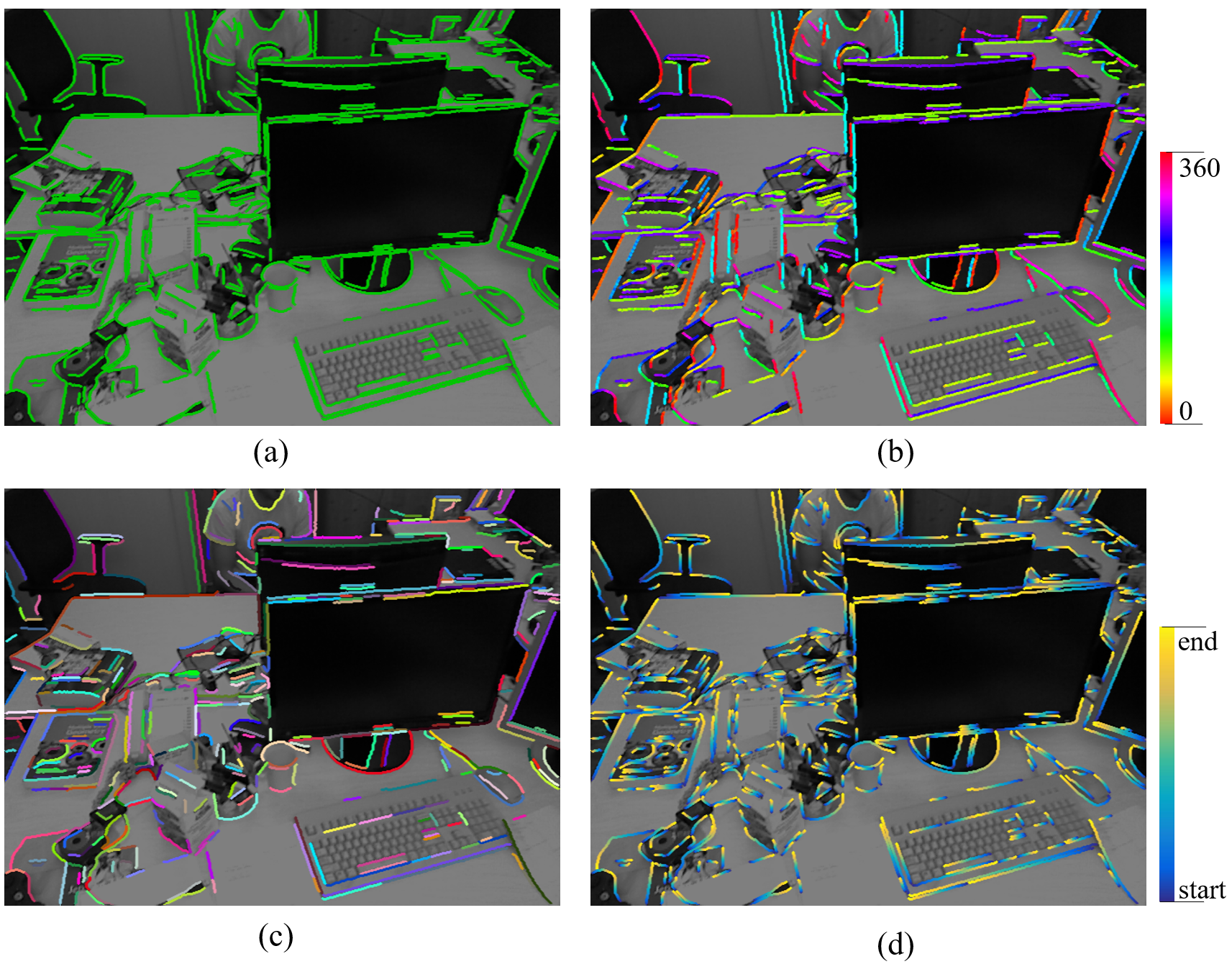}
\caption{An example of organized edge feature extraction from a real-world image. (a) Original Canny output, where extracted edge pixels are shown in green. (b) The image gradient direction of foreground edge pixels. (c) Organized edges in the scene, where each organized edge is assigned a unique color. (d) Sequential information of the organized edges.}
\label{realworld_oedge}
\end{figure}

We employ the Canny algorithm \cite{canny} to extract edges from the image. The non-maximum suppression and double thresholding mechanisms in Canny allow us to extract edges with great coherence and clear structure. A typical edge extraction result obtained from the Canny algorithm is shown in Fig. \ref{extraction_process} (a), where the pixels marked in gray are the detected edges. Additionally, the image gradient direction of each edge pixel can be computed by a Sobel kernel \cite{sobel}, the result is illustrated in Fig. \ref{realworld_oedge} (b).

To extract clustered and sequentialized edges, we treat the edge mask as a topology graph and leverage the Breadth First Search (BFS) algorithm to continuously explore the adjacent pixels of the currently visited inlier pixel to form a connected component. During this exploration process, we actively record the expansion route, which provides the essential information required for the sequentialization of these pixels. Since the non-maximum suppression of the Canny algorithm considers 8 adjacent pixels within a $3 \times 3$ patch around a pixel, we also treat these 8 surrounding pixels as adjacent for BFS expansion to maintain edge consistency. However, performing BFS in an 8-neighborhood configuration can lead to non-unique routes during the expansion on certain edge pixel patterns. This issue arises because the expansion route depends on the order in which BFS explores adjacent pixels, potentially causing ambiguity in the subsequent sequentialization. An example can be found on the lower part of Fig. \ref{extraction_process} (b), where the edge mask on the lower left encounters severe route ambiguity. To address this, we preprocess the edge mask obtained from Canny by suppressing confusing pixel patterns as illustrated in Fig. \ref{extraction_process} (c). We found that path ambiguity only occurs when the two patterns in the upper part of Fig. \ref{extraction_process} (b) appear. To exclude these two patterns, we adjusted their adjacency relationships based on the template shown in Fig. \ref{extraction_process} (b). This preprocessing step ensures that each pixel can preserve a single distinct expansion path during BFS exploration. As illustrated on the lower part of Fig. \ref{extraction_process} (b), after we transformed the left edge mask into the right edge mask through adjacency adjustment, we obtained a concise edge pattern with a single expansion route.

On the assumption of the smoothness of edges, we cluster adjacent edge pixels sharing similar gradient directions (In our implementation, a fixed angular threshold of $20^{\circ}$ is adopted) to formulate an edge feature. By utilizing the modified BFS algorithm to check the smoothness of gradient directions along the edge, we cluster the pixels and simultaneously record the expansion routes of the search. 
Notably, the expansion routes constitute a spanning tree of the search graph, with each pixel maintaining an edge directed to its preceding pixel during traversal. Upon the modified BFS-based traversal and clustering, termed \textit{oriented BFS}, we can acquire a pixel cluster containing the expansion routes of the inlier pixels, as depicted in Fig. \ref{extraction_process} (d). These routes start from the root node of BFS and terminate at the end nodes of the edge. Suppose this cluster is regarded as a directed graph, where the directed edges represent the expansion direction from the current pixel vertex to its expanded pixel vertices. In that case, the end nodes correspond to vertices with zero out-degree, while the root node corresponds to the vertex with zero in-degree. These nodes can be identified effectively by assessing the degree of each pixel in the cluster.

Benefiting from the uniqueness of expansion routes after the preprocessing of the edge mask, we then backtrack along these routes starting from the end nodes to obtain multiple paths pointing from the end node to the root nodes, as illustrated in Fig. \ref{extraction_process} (e). These paths can be regarded as sequential segments that make up the entire edge cluster. The subsequent task then becomes relatively straightforward: to exclude the unnecessary branches and form a complete ordered sequence for the edge cluster, we just need to retain two non-intersecting paths that together form the longest combined path. Then, by reversing the direction of one of these two paths, we ultimately obtain a fully sequentialized edge. The final result of the extraction is shown in Fig. \ref{extraction_process} (f). In summary, we transform the raw edge mask provided by the Canny detector into a set of organized edges. We also provide an example of processing a real image, where Fig. \ref{realworld_oedge} (c) displays the obtained edge clusters, and Fig. \ref{realworld_oedge} (d) illustrates the associated sequential information of these edges. In subsequent sections, we will further discuss how to leverage texture and structural information from these organized edges to facilitate improved association, registration, and joint optimization tasks.

\section{Tracking}
\label{section_tracking}
In this section, we utilize the extracted organized edge features to estimate the transformation between two RGB-D camera frames, which can be formulated as a 3D-2D registration problem. By leveraging the textural and structural information embedded in the organized edges, we propose a tailored tracking approach optimized specifically for edge features. The proposed method employs a coarse-to-fine strategy, effectively balancing efficiency and accuracy.

\subsection{Formulation of 3D-2D Registration}

For a well-aligned RGB-D camera frame, we can typically obtain the depth value $z_i$ of a pixel $\mathbf{p}_i$. Assuming that the camera is fully calibrated, we use a function $\mathbf{x}_i = \pi^{-1}(\mathbf{p}_i, z_i)$ to project $\mathbf{p}_i$ onto the camera coordinate system, with the corresponding inverse function being $\mathbf{p}_i= \pi(\mathbf{x}_i)$. 
We consider a 3D-2D registration problem between a reference frame $\boldsymbol{F}_{r}$ with 3D edge features and the current edge frame $\boldsymbol{F}_{c}$ with 2D edge features. The goal is to estimate the transformation $\mathbf{T} \in \text{SE}(3)$ from $\boldsymbol{F}_{r}$ to $\boldsymbol{F}_{c}$ such that the residual constructed by edges from the two frames is minimized as:
\begin{equation}\label{3d2dalign}
\mathbf{T}^{*} = \arg \min_{\mathbf{T}}\sum_{\mathbf{p}_i\in \boldsymbol{F}_{r}}\Vert{\mathbf{r}_i(\pi^{-1}(\mathbf{p}_i, z_i), \mathbf{T})}\Vert^2
\end{equation}
where the residual $\mathbf{r}_i$ is defined by the associations between  $\boldsymbol{F}_{r}$ and $\boldsymbol{F}_{c}$. Given the varying shapes and lengths of the extracted edge features, directly associating two edges by constructing vectorized texture descriptors is an extremely challenging task. To address this issue, we associate organized edge features by incorporating the similarity of local textural and structural information, as well as the nearest neighbor search. However, when given poor initial poses between the two RGB-D frames, a large amount of nearest neighbor search is required for iterative pose optimization before convergence, which is computationally intensive and time-consuming. Therefore, we adopt a coarse-to-fine tracking strategy. To be specific, we perform coarse tracking to obtain a relatively accurate initial pose firstly. We then perform the edge-wise data associations based on the proper initial pose and refine the pose to achieve more precise tracking using the edge-wise correspondences.

\begin{figure}[h]
\centering
\includegraphics[width=0.9\linewidth]{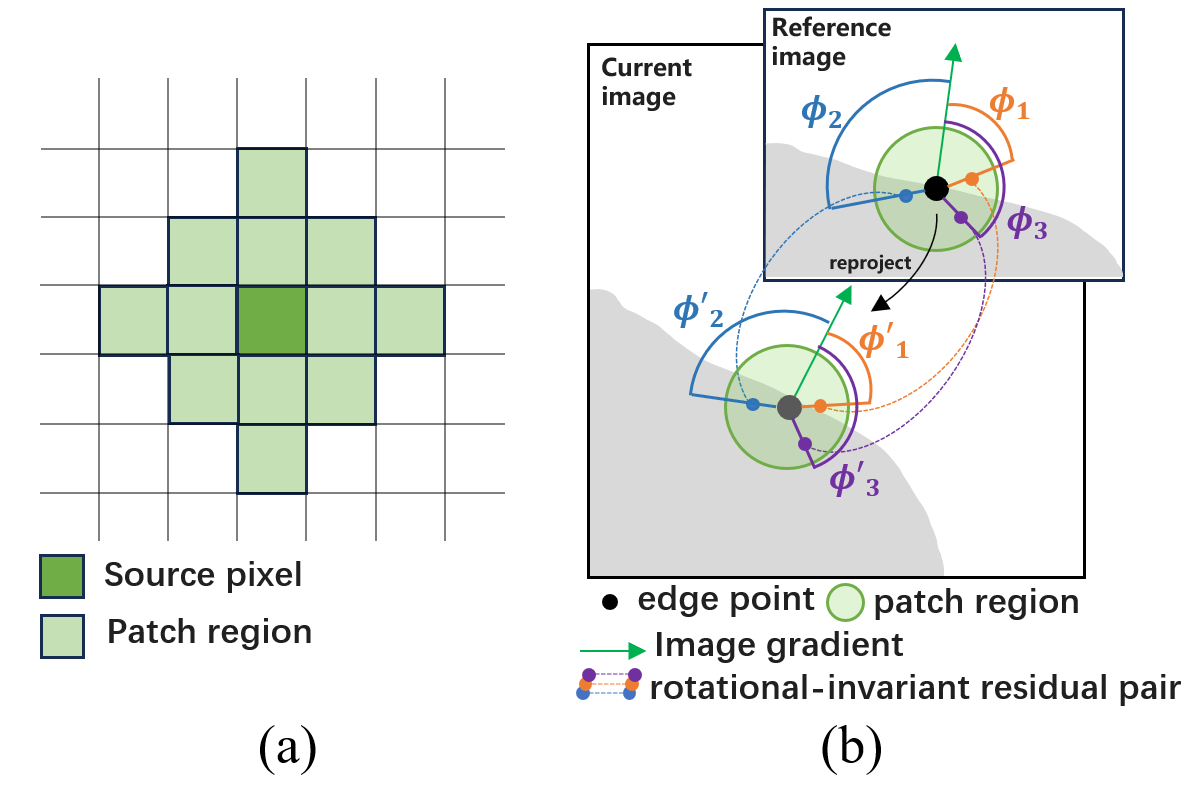}
\caption{An illustration of coarse tracking. (a) A local patch selected for calculating photometric errors. (b) Orientation-dependent photometric residual calculation, where corresponding residual pairs are defined based on the orientations of image gradients.}
\label{figs4f1}
\end{figure}

\subsection{Coarse Tracking Step}
Inspired by~\cite{dso}, to obtain a relatively accurate pose efficiently, we leverage the sparse direct method based on minimizing photometric errors on the extracted edges. Since organized edges have distinct textures, we construct orientation-dependent photometric errors that lead to more deterministic and robust pose estimates.

We denote the grayscale intensity maps of the reference frame and the current frame as $\mathbf{I}_{r}$ and $\mathbf{I}_{c}$, respectively. The general photometric error can be written as $\mathbf{e}_{p}^2 = \Vert \mathbf{I}_{c}[\mathbf{p}'_i]-\mathbf{I}_{r}[\mathbf{p}_i]\Vert^2$, where $\mathbf{p}_i'$ is the re-projection of $\mathbf{p}_i$ given by $\mathbf{p}_i' = \pi( \mathbf{R} \pi^{-1}(\mathbf{p}_i, z_i) + \mathbf{t})$ where $\mathbf{R}\in \text{SO}(3), \mathbf{t} \in \mathbb{R}^3$ represent the camera's 6-DoF pose transformation. To deal with the vagueness and uncertainty of pixel intensities at edges, we construct patch-wise photometric errors in patches as shown in Fig. \ref{figs4f1} (a). The patch coordinates relative to its center are denoted as $\mathcal{O} = \{\mathbf{o}_j\}$. The orientations information of our extracted edge features is found to be very helpful for constructing rotation-invariant photometric errors.  

To perform optimization incorporating the gradient orientations of two edges, we rotate the pixel patch shown in Fig. \ref{figs4f1} (b) to align it with the edge pixel's gradient direction.
For $\mathbf{p}_i$ in $\boldsymbol{F}_{r}$ with the gradient orientation $\theta_i$, as well as its re-projection $\mathbf{p}_i'$ in $\boldsymbol{F}_{c}$ with the gradient orientation $\theta_i'$, the rotation-invariant patch-wise photometric error associated with $\mathbf{p}_i$ is formulated as:
\begin{equation}\label{photometric}
\mathbf{e}_i^2 = w_i\sum_{\mathbf{o}_j\in \mathcal{O}}\Vert \mathbf{I}_{c}[\mathbf{p}'_i+\boldsymbol{\varrho}(\theta_i')\mathbf{o}_j]-\mathbf{I}_{r}[\mathbf{p}_i+\boldsymbol{\varrho}(\theta_i)\mathbf{o}_j]\Vert^2,
\end{equation}
where $\boldsymbol{\varrho}(\cdot)$ is a 2D rotation on the image plane with the rotation angle given by the image gradient $\theta_i$ or $\theta_i'$, and $w_i$ is an adaptive weight for robust optimization, defined as:
\begin{equation}
\label{coarseWeight}
w_i= \frac{w_s(\mathbf{p}_i)}{1+\Vert\nabla \mathbf{I}_{r}(p_i)\Vert^2/255^2},
\end{equation}
where $w_s(p_i)\in[0,1]$ is a hyperparameter reflecting the quality of sensor observation, its form will be discussed in detail in Section \ref{config_study}.
This weighting scheme down-weights edges with high image gradients, as the goal is to align all extracted edges effectively. 
%
To achieve high efficiency in coarse tracking, edge pixels are uniformly sampled from the organized edge features to construct photometric errors.

\begin{figure}[t]
\centering
\includegraphics[width=3 in]{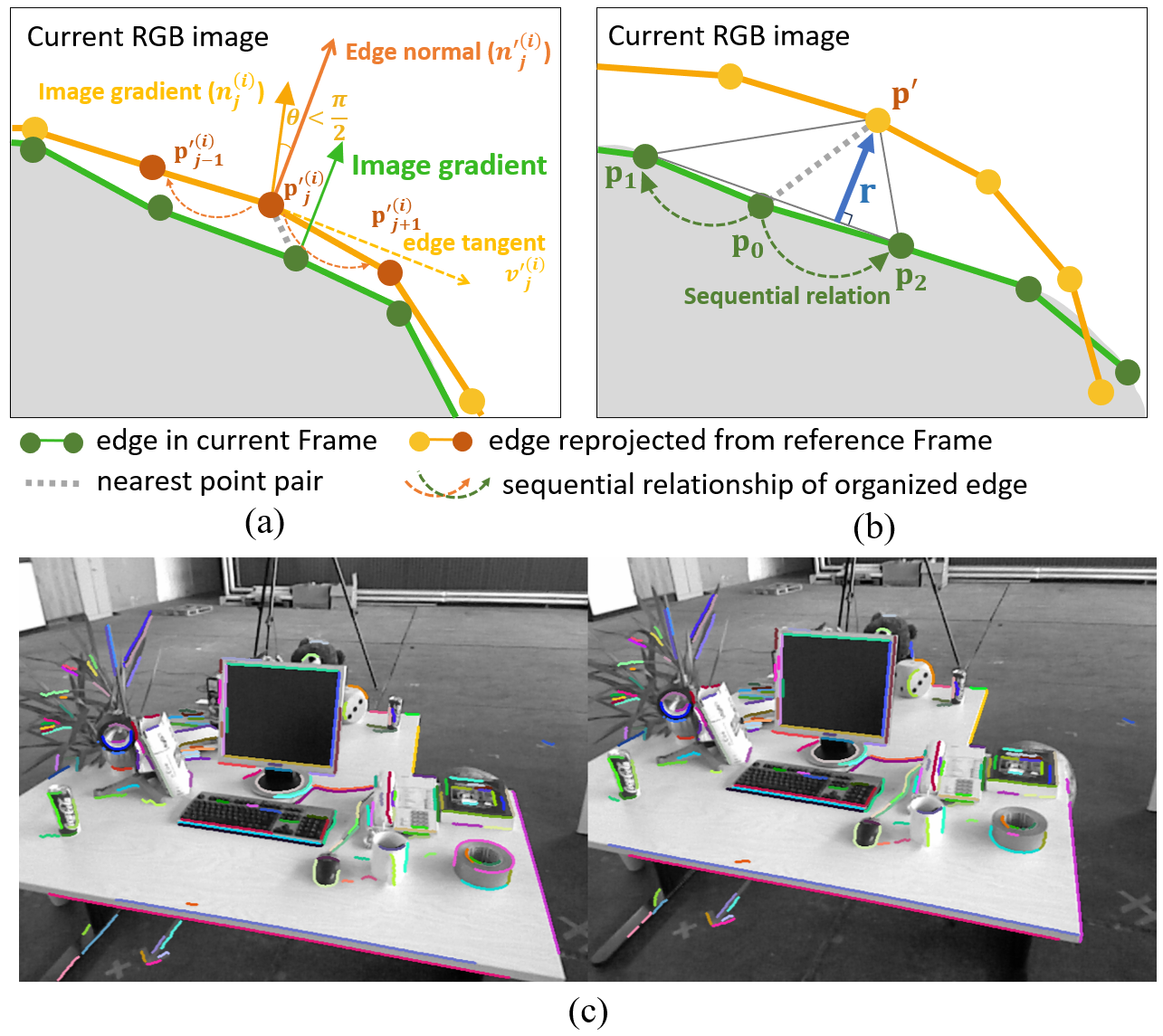}
\caption{An illustration of fine tracking and edge-wise association. (a) Details of edge association. We calculate the normals of the reprojected edge(orange) and the image gradients of the current frame's edge(green), thereby associating edges with constant structural and textual characteristics. (b) Illustration of the point-to-tangent residual, where the final residual $r$ is calculated through a triangle formed by $p_0,p_1,p_2$. (c) An example of edge-wise association between two image frames that have a certain viewpoint difference, where associated edges are rendered with the same color.}
\label{fine_track}
\end{figure}

\subsection{Fine Tracking Step}\label{fine}
Given a relatively accurate pose prior obtained from coarse tracking, we can perform reliable edge-wise data association for fine tracking. After the feature extraction, an RGB-D frame contains a set of the extracted organized edges, denoted by $\mathcal{C} = \{\mathcal{E}_i\}$, where each organized edge is a collection of sequentialized points, denoted as $\mathcal{E}_i = \{\mathbf{p}_j^{(i)}\}$. We denote the edge set of $\boldsymbol{F}_{r}$ as $\mathcal{C}_{r}$, and $\mathcal{C}_{c}$ of $\boldsymbol{F}_{c}$, respectively.

To establish associations between $\mathcal{C}_{r}$ and $\mathcal{C}_{c}$, taking $\mathcal{E}_i \in \mathcal{C}_{r} $ as an example, we construct a 2D KD-tree for all edge feature points in $\mathcal{C}_{c}$. Then, $\mathcal{E}_i$ is projected onto $\boldsymbol{F}_{c}$ using the transformation obtained by coarse tracking. The projected edge is denoted as $\mathcal{E}_i'$, where each edge point represented as ${\mathbf{p}'}_j^{(i)}$. Next, a radius neighbor search is performed around ${\mathbf{p}'}_j^{(i)}$ to identify candidate edge points in $\mathcal{C}_{c}$. Then, these candidates are further validated using local textural and structural information, which is represented as an edge normal-based association.

As shown in Fig. \ref{fine_track} (a), based on the sequential relationships of organized edges, we retrieve the preceding point ${\mathbf{p}'}_{j-1}^{(i)}$ and the succeeding point ${\mathbf{p}'}_{j+1}^{(i)}$ of current point ${\mathbf{p}'}_j^{(i)}$ and define the tangent vector of ${\mathbf{p}'}_j^{(i)}$ as ${\mathbf{v}'}_j^{(i)}={\mathbf{p}'}_{j+1}^{(i)}-{\mathbf{p}'}_{j-1}^{(i)}$. The normalized vector orthogonal to ${\mathbf{v}'}_j^{(i)}$ is then derived as:
\begin{equation}
\label{orthogonal}
{\mathbf{m}'}_j^{(i)} = \frac{1}{\Vert{\mathbf{v}'}_{j}^{(i)}\Vert}
\begin{bmatrix}
0 & -1 \\ 1 & 0
\end{bmatrix}{\mathbf{v}'}_{j}^{(i)}.
\end{equation}
The normal vector ${\mathbf{n}'}_j^{(i)}$ of ${\mathbf{p}'}_j^{(i)}$ is then defined as the vector orthogonal to the edge and in the direction consistent with the image gradient from the original image, formulated as:
\begin{equation}
\label{normal}
{\mathbf{n}'}_j^{(i)} = \text{sgn}({\mathbf{m}'}_{j}^{(i)}, \nabla \mathbf{I}_{r}[\mathbf{p}_j^{(i)}]){\mathbf{m}'}_{j}^{(i)},
\end{equation}
where $\text{sgn}(a,b)$ is a sign function based on the inner product of vectors $a$ and $b$, given by:
\begin{equation}\label{sign}
\text{sgn}(\boldsymbol{a},\boldsymbol{b})= 
\left\{\begin{aligned}
-1 \quad &\text{if} \quad \boldsymbol{a}^{\mathsf{T}}\boldsymbol{b} < 0,\\
1 \quad &\text{if} \quad \boldsymbol{a}^\mathsf{T}\boldsymbol{b} \geq 0.
\end{aligned}\right.
\end{equation}
To determine whether ${\mathbf{p}'}_j^{(i)}$ can be associated with its neighboring candidate points, we check whether the direction of the normal vector ${\mathbf{n}'}_j^{(i)}$ is consistent with those of the normals of these candidate points. Assuming that the normal vectors of edge points in $\boldsymbol{F}_{c}$ are represented by the image gradient, we then calculate the angle between ${\mathbf{n}'}_j^{(i)}$ and the image gradients of the candidate pixels to determine whether those candidates can be associated with ${\mathbf{p}}_j^{(i)}$, Our experiments demonstrate that setting an exceptionally strict angular threshold of $5^{\circ}$ yields highly robust and precise feature associations. We therefore adopt this threshold value as a fixed configuration parameter.

After establishing the point-wise correspondence for each ${\mathbf{p}}_j^{(i)}\in \mathcal{E}_i$, we further obtain the edge-wise association for $\mathcal{E}_i$ through a voting-based method: when the points associated with $\mathcal{E}_i$ in $\mathcal{E}_j \in \mathcal{C}_{c} $ exceed a certain proportion, we consider $\mathcal{E}_j$ to be associated with $\mathcal{E}_i$. Since edges may be fragmented during clustering, one-to-many edge associations between $\mathcal{E}_i$ in $\mathcal{C}_{r}$ and $\mathcal{E}_j$ in $\mathcal{C}_{c}$ are allowed. We further cluster the matched edges that are associated with an identical edge in the other frame. The final associated and clustered edges are showcased in Fig. \ref{fine_track} (c).

Through the association process, edge-wise associations are established based on the edge normals, since some of the edges in $\mathcal{C}_{r}$ may fail to establish correspondence with edges in $\mathcal{C}_{c}$, edge points $p_j^{(i)}\in \mathcal{C}_{r}$ can be categorized into two sets: the ones with correspondences denoted as $\mathcal{P}_{a}$ and others without correspondence denoted as $\mathcal{P}_{r}$. For $\mathcal{P}_{a}$, we minimize both the photometric errors and the geometric errors for pose estimation. In contrast,  for $\mathcal{P}_{r}$, only photometric errors are minimized.

The process of geometric residual construction is illustrated in Fig. \ref{fine_track} (b). For $\mathbf{p} \in \mathcal{P}_{a}$,  we denote its reprojected position in $\boldsymbol{F}_{c}$ as $\mathbf{p}'$, and its associated point in $\boldsymbol{F}_{c}$ as $\mathbf{p}_0$. Based on the sequential relationship of the organized edges, we retrieve the preceding point $\mathbf{p}_1$ and the succeeding point $\mathbf{p}_2$ of $\mathbf{p}_0$ to determine its tangent vector as $\mathbf{p}_2-\mathbf{p}_1$, then we construct a point-to-tangent geometric residual $\mathbf{e}_{g} \in \mathbb{R}^2$, with:
\begin{equation}
\label{e_ego}
\mathbf{e}_{g} = (\mathbf{p}'-\mathbf{p}_1)-\frac{\mathbf{p}_2 - \mathbf{p}_1}{\Vert \mathbf{p}_2-\mathbf{p}_1\Vert}\left[\frac{(\mathbf{p}_2 - \mathbf{p}_1)^T}{\Vert \mathbf{p}_2-\mathbf{p}_1 \Vert}(\mathbf{p}'-\mathbf{p}_1) \right].
\end{equation}
The Jacobian with respect to the 6-DoF pose can be derived by applying the left perturbation model with $ \delta\bm{\xi} \in \mathfrak{se}(3)$, given by:
\begin{equation}
\label{fine_jacobian}
\mathbf{J}_{g} = \frac{\partial \mathbf{e}_{g}}{\partial \delta\bm{\xi}} = \left(\mathbf{I}_{3\times 3} - \frac{(\mathbf{p}_2 - \mathbf{p}_1)(\mathbf{p}_2 - \mathbf{p}_1)^\mathsf{T}}{\Vert \mathbf{p}_2-\mathbf{p}_1\Vert^2} \right)\frac{\partial \mathbf{p}'(\bm{\xi})}{\partial \delta\bm{\xi}}.
\end{equation}
where $\partial \mathbf{p}'(\bm{\xi})/\partial \delta\bm{\xi}$ represents the Jacobian of the 3D-2D projection with its analytical form provided in Appendix \ref{app_j}. For $\mathcal{P}_{a}$, both geometric and photometric residuals are taken into account, expressed as:
\begin{equation}
\label{e_asso}
\mathbf{e}_{a} = 
\begin{bmatrix}
\mathbf{e}_{g} \\ \alpha \mathbf{e}_{p}
\end{bmatrix},
\mathbf{J}_{a} =
\begin{bmatrix}
\mathbf{J}_{g} \\ \alpha \mathbf{J}_{p}
\end{bmatrix}
\end{equation}
where $\mathbf{J}_{p}$ is the Jacobian matrix based on the photometric error described in Equation (\ref{photometric}), and $\alpha$ is a scaling factor used to balance the magnitudes of the geometric and photometric residuals. For $\mathcal{P}_{r}$, we compute $\mathbf{J}_{r} = \alpha\mathbf{J}_{p}$ as the Jacobian matrix of the photometric residual. Thus, the optimization  process is formulated as a nonlinear least squares problem:
\begin{equation}
\label{s4_optimize}
\bm{\xi}^* = \arg\min_{\bm{\xi}}(\sum_{\mathbf{p}^{(i)}\in\mathcal{P}_{a}}(\mathbf{e}^{(i)}_{a}(\bm{\xi}))^2 + \sum_{\mathbf{p}^{(j)}\in\mathcal{P}_{r}}(\mathbf{e}^{(j)}_{r}(\bm{\xi}))^2)
\end{equation}
where $\mathbf{e}_r$ is the photometric error described in Equation (\ref{photometric}). We employ the Gauss-Newton method to optimize $\bm{\xi}$ iteratively, where increment $\Delta\bm{\xi}$ at each iteration is given by:
\begin{equation}
\label{s4_gn}
\Delta\bm{\xi} = -\mathbf{H}^{-1}(\bm{\xi})g(\bm{\xi})
\end{equation}
where $\mathbf{H}$ is defined as:
\begin{equation}
\label{s4_H}
\mathbf{H}=\sum_{\mathbf{p}^{(i)}\in\mathcal{P}_{a}}(\mathbf{J}_{g}^\mathsf{T}\mathbf{J}_{g}+\mathbf{J}_{p}^\mathsf{T}\mathbf{J}_{p})^{(i)} + \sum_{\mathbf{p}^{(j)}\in\mathcal{P}_{r}}(\mathbf{J}_{p}^\mathsf{T}\mathbf{J}_{p})^{(j)}
\end{equation}
and $\mathbf{g}$ is defined as:
\begin{equation}
\label{s4_g}
\mathbf{g} =\sum_{\mathbf{p}^{(i)}\in\mathcal{P}_{a}}(\mathbf{J}_{a}^\mathsf{T}\mathbf{e}_{a})^{(i)} + \sum_{\mathbf{p}^{(j)}\in\mathcal{P}_{r}}(\mathbf{J}_{r}^\mathsf{T}\mathbf{e}_{r})^{(j)}
\end{equation}

During the tracking step, a new keyframe is inserted when the ratio of associable organized edge features between the current frame and the reference frame falls below 80\%. To achieve a more robust optimization, we also employ the $\chi^2$ test\cite{robust_wight_dvo, orb-slam3} and exclude residual pairs with $\chi^2$ values exceeding the threshold to mitigate the impact of inaccurate observations on the overall optimization outcome.

\section{Local Mapping}\label{section_joint}
Organized edge features can establish edge-wise associations, enabling us to perform bundle adjustment (BA) to jointly optimize the 3D structures and poses across multiple frames. 

\subsection{Co-visibility Graph}\label{section_co}
\begin{figure}[ht]
\centering
\includegraphics[width=0.9\linewidth]{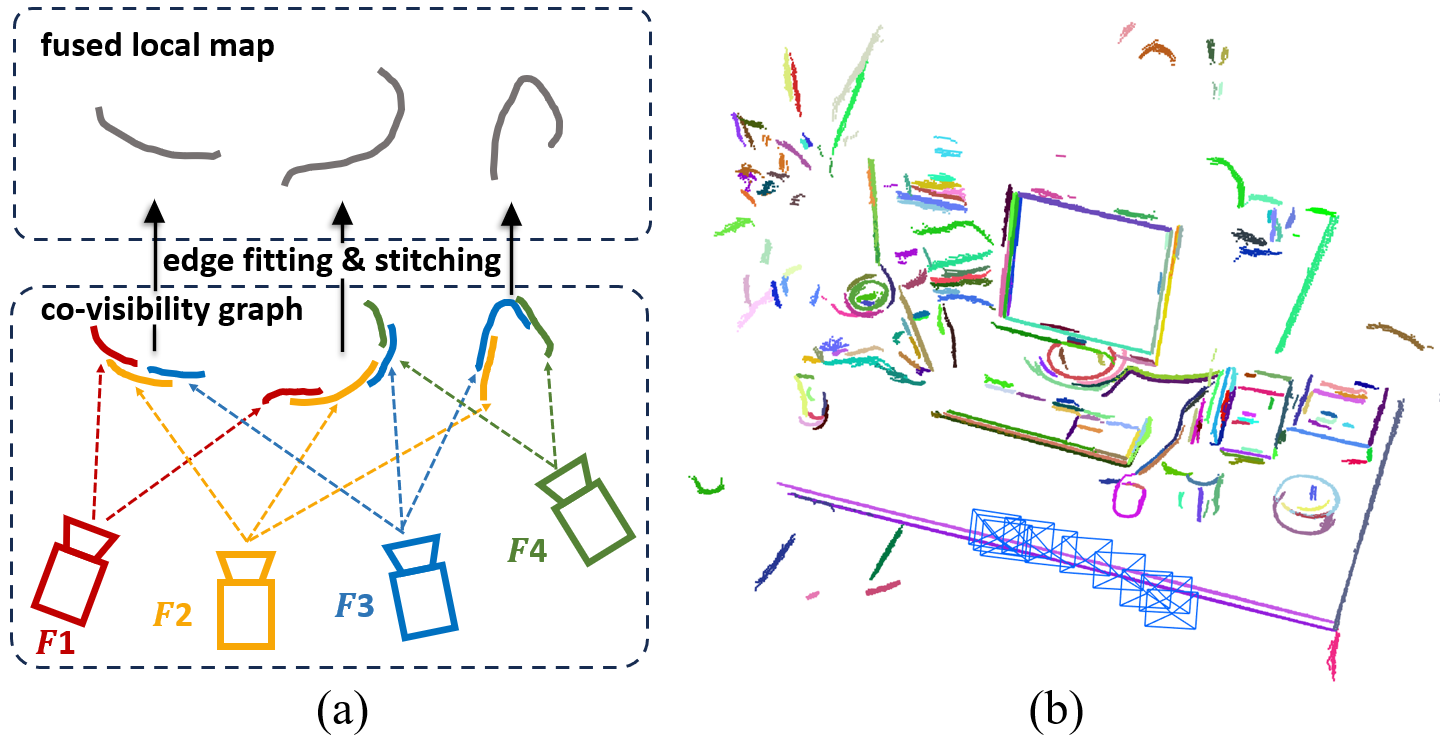}
\caption{(a) Illustration of a co-visibility graph for organized edges, along with the local map built upon the co-visibility graph.  (b) An example of a co-visibility graph containing several keyframes in a real-world environment\cite{tum}, where edges belonging to the same cluster are rendered by the same color. }
\label{co_visibility}
\end{figure}

 To perform BA, the first step is to establish a co-visibility graph between the edge structures and involved frames, as illustrated in Fig. \ref{co_visibility} (a).  While the method described in Section \ref{fine} allows us to associate edges between two frames, to further link organized edges across multiple frames, we employ a disjoint set \cite{disjoint} to facilitate associations across multiple frames. The disjoint set is an efficient data structure that clusters elements by maintaining a tree-like membership. This makes it particularly well-suited for managing associations in multi-frame scenarios.

For a set of keyframes $\{\boldsymbol{F}_1,\boldsymbol{F}_2,\cdots,\boldsymbol{F}_{i-1}\}$ with an established co-visibility graph represented by a disjoint set, inserting a new keyframe $\boldsymbol{F}_i$ into the graph requires sequentially associating $\boldsymbol{F}_i$ with each $\boldsymbol{F}_j,(j<i)$ while continuously updating the disjoint set. To enhance the efficiency, we start establishing association from $\boldsymbol{F}_{i-1}$, as $\boldsymbol{F}_{i-1}$ are most likely to share the strongest co-visibility relationship with $\boldsymbol{F}_i$. In addition, once an edge is associated with an existing edge from the co-visibility graph, it does not need to be re-associated with others. The specific algorithm of updating the co-visibility graph by a disjoint set is outlined in the Algorithm \ref{algorithm_update}. We also provide the algorithm for initializing a co-visibility graph with the disjoint set in the Appendix \ref{app_alg}. A resulting co-visibility graph in a real-world environment are shown in Fig. \ref{co_visibility} (b).

 \begin{algorithm}[ht]
\caption{Update a Co-visibility Graph }\label{algorithm_update}
\begin{algorithmic}
\STATE \textbf{Input} Current keyframe $\boldsymbol{F}_{c}$ with included edges $\{ ^{c}\mathcal{E}_1, ^{c}\mathcal{E}_2, \cdots , ^{c}\mathcal{E}_N\}$; disjoint set of the current local map $\mathcal{D}$; keyframe list of the current local map $\mathcal{K} = \{\boldsymbol{F}_1, \cdots, \boldsymbol{F}_M\}$
\STATE \textbf{Output} updated disjoint set $\mathcal{D}$, updated keyframe stack $\mathcal{K}$.
\STATE 
\STATE associate\_list = \{\textbf{false},\textbf{false},...,\textbf{false}\}(size = $N$)

\STATE \textbf{for} $\boldsymbol{F}_i \in \mathcal{K}$, $i = M, M-1, ..., 1$, \textbf{do:}
\STATE \hspace{0.5cm}\textbf{for} $^{c}\mathcal{E}_j \in \boldsymbol{F}_{c}$, $j = 0, 1, ..., N$, \textbf{do:}
\STATE \hspace{1.0cm} \textbf{if} associate\_list[$j$] $\neq $ \textbf{false}, \textbf{do:}
\STATE \hspace{1.5cm} \textbf{continue}
\STATE \hspace{1.0cm} $\{^{i}\mathcal{E}_{a_1}, ..., ^{i}\mathcal{E}_{a_n}\} \gets $\textbf{edge\_association}($^{cur}\mathcal{E}_j$, $\boldsymbol{F}_i$)
\STATE \hspace{1.0cm} \textit{tuple} $\boldsymbol{e}_1 \gets (\boldsymbol{F}_i.\mathbf{id}, a_1)$  
\STATE \hspace{1.0cm} \textbf{for} $^{i}\mathcal{E}_{a_m} \in \{^{i}\mathcal{E}_{a_1}, ..., ^{i}\mathcal{E}_{a_n}\}$, $m\neq 1$ \textbf{do:}
\STATE \hspace{1.5cm} \textit{tuple} $\boldsymbol{e}_2 \gets (\boldsymbol{F}_i.\mathbf{id}, a_m)$
\STATE \hspace{1.5cm} $\mathcal{D}$.\textbf{locate}($\boldsymbol{e}_1$).\textbf{set\_union}( $\mathcal{D}$.\textbf{locate}($\boldsymbol{e}_2$))
\STATE \hspace{1.0cm} $A \gets \{^{i}\mathcal{E}_{a_1}, ..., ^{i}\mathcal{E}_{a_n}\}$
\STATE \hspace{1.0cm} \textit{tuple} $\boldsymbol{e}_{new} \gets (\mathcal{F}_{cur}, j)$
\STATE \hspace{1.0cm} $\mathcal{D}$.\textbf{insert}($\boldsymbol{e}_{new}$)
\STATE \hspace{1.0cm} \textbf{if} $A \neq \varnothing$, \textbf{do:}
\STATE \hspace{1.5cm} $\mathcal{D}$.\textbf{locate}($\boldsymbol{e}_{new}$).\textbf{set\_union}( $\mathcal{D}$.\textbf{locate}($\boldsymbol{e}_1$))
\STATE \hspace{1.5cm} associate\_list[$j$] $\gets$ \textbf{true}
\STATE \textbf{for} $^{cur}\mathcal{E}_j \in \boldsymbol{F}_{c}$, $j = 0, 1, ..., N$, \textbf{do:}
\STATE \hspace{0.5cm} \textbf{if} associate\_list[$j$] $\neq $ \textbf{true}, \textbf{do:}
\STATE \hspace{1.0cm} \textit{tuple} $\boldsymbol{e}_{new} \gets (\boldsymbol{F}_{c}, j)$
\STATE \hspace{1.0cm} $\mathcal{D}$.\textbf{insert}($\boldsymbol{e}_{new}$)
\STATE $\mathcal{K}$.\textbf{push}($\boldsymbol{F}_{c}$)
\STATE \textbf{return} $\mathcal{K}$, $\mathcal{D}$

\end{algorithmic}
\label{alg0}
\end{algorithm}

\subsection{Edge Fusion}\label{edge_fu}
After establishing the co-visibility graph, the next step is to merge the associated multiple edges across multiple keyframes into a single unified edge that serves as a map element for local mapping. Fusing a group of edges involves two main steps: \textit{fitting} and \textit{stitching}. This two-step paradigm arises from the partial observability of spatial edges from different viewpoints. Based on the sequential information inherent in organized edges, we can seamlessly perform both fitting and stitching without significantly distorting the inherent edge shapes.

\begin{figure}[ht]
\centering
\includegraphics[width=0.9\linewidth]{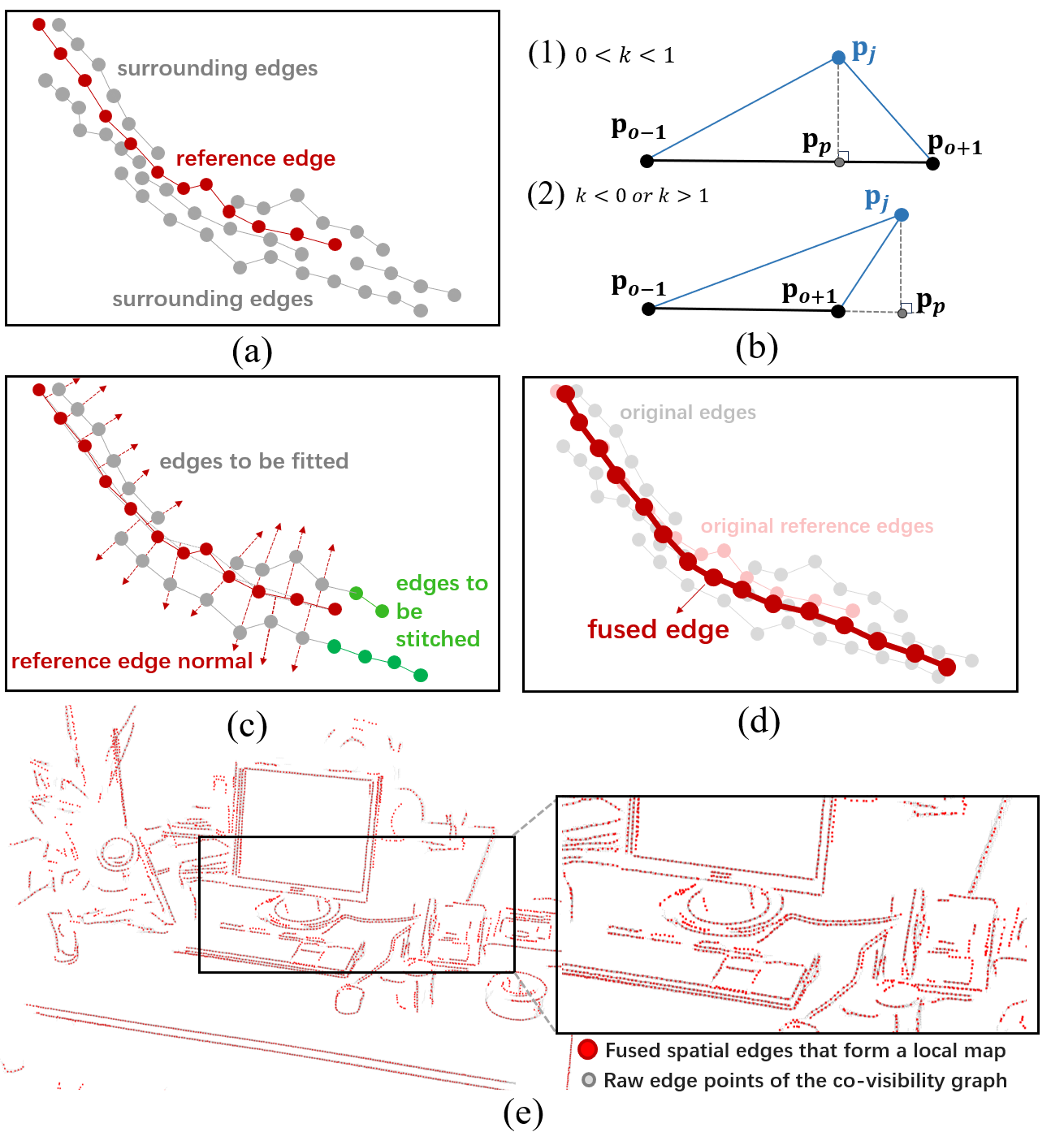}
\caption{Illustrations of the edge fusing process. (a) A cluster of edges that need to be merged.  (b) Two cases of the foot of perpendicular lines. (c) Process of edge normal-based merging, where gray segments can be fitted with the reference edge, and green segments need to be stitched. (d) Result of edge fusion. (e) A fused local map created by edge fitting and stitching, with fully preserved shapes of edge features.}
\label{edge_fusion}
\end{figure}

For edge fitting, we first back-project the edges into 3D space, where a set of clustered edges from the co-visibility graph is illustrated in Fig.\ref{edge_fusion} (a). We then perform edge fitting and stitching with the assist of edge normals. First, we select the longest edge among the cluster as the reference edge, denoted as $\boldsymbol{\mathcal{E}}_{r}$, with its 3D edge points $\{^{r}\mathbf{p}_i\}$. For an edge point $\mathbf{p}_j$ from the neighboring edges, we perform a nearest neighbor search to find its closest reference edge point, denoted as $^{r}\mathbf{p}_o$, and then retrieve its preceding point $^{r}\mathbf{p}_{o-1}$ and its succeeding point $ ^{r}\mathbf{p}_{o+1}$. Next, we calculate the foot of the perpendicular line from $\mathbf{p}_j$ to the line formed by $^{r}\mathbf{p}_{o-1}$ and $ ^{r}\mathbf{p}_{o+1}$, denoted as $\mathbf{p}_p$. To determine whether $\mathbf{p}_j$ can fit with $\boldsymbol{\mathcal{E}}_{r}$ in the normal direction, we formulate a criterion $k$ as:
\begin{equation}
\label{s5_k}
k = -\frac{(^{r}\mathbf{p}_{o-1} - \mathbf{p}_j)^\mathsf{T}(^{r}\mathbf{p}_{o+1} - ^{r}\mathbf{p}_{o-1})}{\Vert^{r}\mathbf{p}_{o+1} - ^{r}\mathbf{p}_{o-1}\Vert^2}.
\end{equation}
For $0<k<1$, it corresponds to case (1) shown in Fig. \ref{edge_fusion} (b), where the perpendicular foot falls between $\mathbf{p}_{o-1}$ and $\mathbf{p}_{o+1}$. And for $k \leq 0$ or $k \geq 1$, it corresponds to case (2) where the perpendicular foot falls outside. We conclude that $\mathbf{p}_j$ can fit with $\boldsymbol{\mathcal{E}}_{r}$ in the normal direction only when $0<k<1$, Accordingly, the foot of the perpendicular $\mathbf{p}_p$ is derived by
\begin{equation}
\label{s5_pp}
\mathbf{p}_p = k(^{r}\mathbf{p}_{o+1} - ^{r}\mathbf{p}_{o-1}) + ^{r}\mathbf{p}_{o-1}.
\end{equation}
During fitting, we associate $\mathbf{p}_j$ with the nearest point among $^{r}\mathbf{p}_{o-1},^{r}\mathbf{p}_{o}$ and $^{r}\mathbf{p}_{o+1}$ that has the smallest distance to $P_p$.As a result, we obtain a set of points $\{\mathbf{p}_j\}$ for each reference edge point $^{r}\mathbf{p}_i \in \boldsymbol{\mathcal{E}}_{r} $ associated with it in the direction of edge normal.

After the classification based on the two cases given by Eq. \ref{s5_k}, all of the points on surrounding edges are classified as ones that can be fitted with $\boldsymbol{\mathcal{E}}_{r}$ and the others that need to be stitched, as shown in Fig. \ref{edge_fusion} (c). Points that can be fitted are used for fitting a merged edge feature $\bar{\boldsymbol{\mathcal{E}}}_{r}$ in the local map. While the rest are stitched into $\boldsymbol{\mathcal{E}}_{r}$. When a new segment is stitched into $\boldsymbol{\mathcal{E}}_{r}$, it will be considered as a part of $\boldsymbol{\mathcal{E}}_{r}$ to continue merging preceding associated edge points.

To compute a fitted edge point $\bar{\mathbf{p}}$ on the fused edge feature $\bar{\boldsymbol{\mathcal{E}}}_{r}$, we simultaneously calculate the centroid of the associated points and its covariance $\mathbf{A}$ as follows:
\begin{equation}
\label{s5_p_bar}
\bar{\mathbf{p}} = \frac{1}{N}\sum_{j=1}^N \mathbf{p}_j,\; A=\frac{1}{N}\sum_{j=1}^N (\mathbf{p}_j-\bar{\mathbf{p}})(\mathbf{p}_j-\bar{\mathbf{p}})^\mathsf{T}.
\end{equation}

A sharp edge feature is supposed to have concentrated edge points, corresponding to small eigenvalues of $\mathbf{A}$. We use $\lambda(\mathbf{A})$  to represent the largest eigenvalue of the covariance matrix $\mathbf{A}$. When $\lambda(\mathbf{A})$ of the associated points is large, it usually means edges within the edge cluster are spaced far apart, typically corresponding to edges that are not actual spatial features but are artifacts of the camera viewpoint or dynamic disturbances. 
Therefore, we omit the edges with large $\lambda(\mathbf{A})$. Ultimately, we obtain the fitted edges for each edge cluster, as shown in Fig. \ref{edge_fusion} (d). The final processing results in a real-world environment are shown in Fig. \ref{edge_fusion} (e), where all edges have undergone high-quality fitting while preserving their original shapes.

\subsection{Bundle Adjustment}\label{ba}
Based on the co-visibility graph and fused edge features in the local map, we can proceed with bundle adjustment to jointly optimize the spatial edge features and poses of multiple keyframes. Given a series of poses $\boldsymbol{T}=\{\mathbf{T}_1, \mathbf{T}_2,\cdots, \mathbf{T}_M\}$ to be jointly optimized on the local map, along with a series of fused edges $\boldsymbol{E} = \{\bar{\boldsymbol{\mathcal{E}}}_1, \bar{\boldsymbol{\mathcal{E}}}_2,\cdots, \bar{\boldsymbol{\mathcal{E}}}_N\}$, the bundle adjustment problem can be formulated as:
\begin{equation}
\label{s5_ba_total}
(\boldsymbol{T}^*, \boldsymbol{E}^*) = \arg\min_{\boldsymbol{T,E}}\sum_{i=1}^M\sum_{j=1}^N \Vert \boldsymbol{l}(^i\mathcal{E}_{sj}, \mathbf{T}_i^{-1}\bar{\boldsymbol{\mathcal{E}}}_j)\Vert^2_\Sigma.
\end{equation}
where $^i\mathcal{E}_{sj}$ denotes the image edge associated with the local map edge $\bar{\boldsymbol{\mathcal{E}}}_j$ in frame $\boldsymbol{F}_i$, and $ \boldsymbol{l}(^i\mathcal{E}_{sj},  \mathbf{T}_i^{-1}\bar{\boldsymbol{\mathcal{E}}}_j)$ represents the geometric residual given by Eq. \ref{e_ego} between the image edge and its corresponding local map edge. Since organized edge features do not align at the pixel level, we decouple the bundle adjustment task, inspired by \cite{balm}, which turns Equation (\ref{s5_ba_total}) into:
\begin{equation}
\label{s5_ba_decouple}
(\boldsymbol{T}^*, \boldsymbol{E}^*) = \arg\min_{\boldsymbol{T}} (\underbrace{\arg\min_{\boldsymbol{E}}\sum_{i=1}^M\sum_{j=1}^N \Vert  \boldsymbol{l}(^i\mathcal{E}_{sj},  \mathbf{T}_i^{-1}\bar{\boldsymbol{\mathcal{E}}}_j)\Vert^2_\Sigma)}_{\text{edge fitting}}.
\end{equation}
We find that this decoupling splits the original BA into two tasks: fitting and pose-only optimization. In the fitting stage, edges are fitted into analytical forms that minimizes the geometric residual related to $\boldsymbol{T}$, and then the pose-only optimization is conduct to optimize all the poses $\boldsymbol{T}=\{ \mathbf{T}_1, \mathbf{T}_2,\cdots, \mathbf{T}_M\}$. This decoupling method transforms the BA problem into a clearer form when the pixel-wise reprojection error is difficult to construct, but the geometric residual of features is easy to define. However, edges with diverse shapes do not have an analytical formulation for fitting. Therefore, to facilitate the construction of optimization, we directly apply the edge fusion method described in Section \ref{edge_fu} as the fitting results. Since the fused edges are formed by the centroids obtained by associating multiple edges along the normal direction, they approximately preserve the minimum fitting geometric residuals. Then, the subsequent pose-only optimization problem is no longer a joint optimization problem for $\boldsymbol{T}$, but is decomposed into independent problems for each $\mathbf{T}_i \in \boldsymbol{T}$. Specifically, for each $\mathbf{T}_i$, the simplified optimization problem is given by:
\begin{equation}
\label{s5_align}
\mathbf{T}_i^* = \arg\min_{\mathbf{T}_i}\sum_{j=1}^N\Vert \boldsymbol{l}(^i\mathcal{E}_{sj}, \mathbf{T}_i^{-1}\bar{\boldsymbol{\mathcal{E}}}_j)\Vert^2_\Sigma.
\end{equation}
This optimization task can be viewed as aligning the local map $\boldsymbol{\mathcal{E}}$ with each camera pose $\mathbf{T}_i\in\boldsymbol{T}$. Following the normal-based tracking procedure discussed in Section \ref{fine}, we cast this optimization problem as a 3D-2D registration problem. Based on these formulations, we alternately perform edge fusion and registration to obtain a converged result. At the same time, considering the computational efficiency, we employ a sliding window mechanism to continuously update the local map in real-time as the camera stream progresses.

\section{Experiments}\label{expr}

We implemented the proposed VO system in C++ and evaluated its performance in terms of accuracy, robustness, and computational efficiency. The full pipeline was tested in different configurations against state-of-the-art algorithms across various datasets, ranging from publicly available real-world benchmarks, synthetic benchmarks, and self-collected sequences in real indoor and outdoor scenarios. A comprehensive study was conducted to assess the performance of the proposed system in both general indoor scenes and specialized indoor and outdoor environments characterized by distinct structures and textures.

\subsection{Configuration Studies}\label{config_study}

\begin{figure}[h]
\centering
\includegraphics[width=0.9\linewidth]{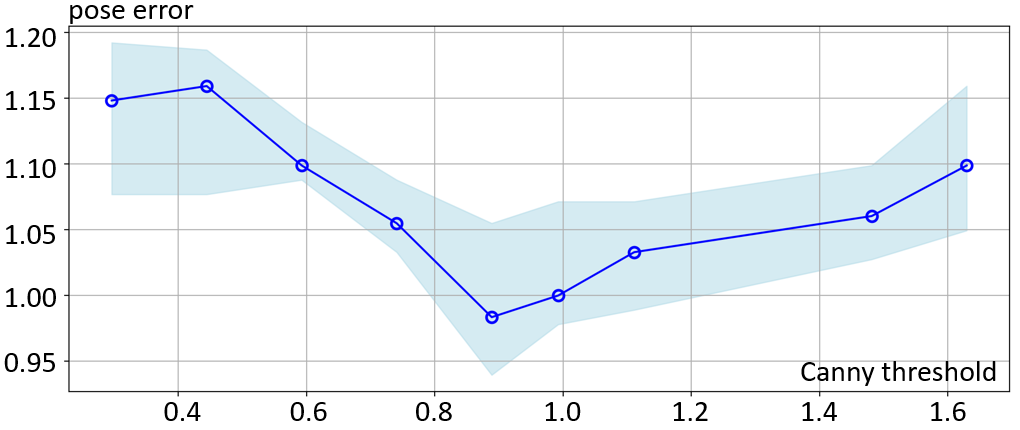}
\caption{Trajectory errors under different Canny thresholds (scaled based on the reference threshold).}
\label{thres}
\end{figure}

In this subsection, we explore the optimal configuration for localization performance. First, we discuss the accurate extraction of features. We investigated the impact of different thresholds of the Canny edge detector on the results. Canny uses a double threshold approach, where the high threshold primarily influences the number of edges extracted, while the low threshold mainly affects the coherence of the extracted edges. Due to the high-quality edge clustering and merging, our odometry system is robust to the incoherence of extracted edges. Hence, we focus mainly on the selection of the high threshold. We evaluated the trajectory errors under varying high thresholds of the Canny detector, with the low threshold kept as half of the high threshold. Given the variability of image sequences, we used the Otsu thresholding algorithm \cite{otsu} to binarize the gradient images obtained from the Sobel kernel. The resulting threshold was then used as a reference value. For comparison, both trajectory errors and Canny thresholds were scaled according to this reference threshold and presented in Fig. \ref{thres}.

We found that as the Canny threshold gradually increases, the trajectory error decreases first and then starts to increase. The best results are achieved when the threshold is close to the reference value. Larger thresholds tend to result in a deficiency of adequate repeatable edge features for association and localization, while smaller thresholds extract a lot of image noise as edges. Therefore, we conclude that selecting the Canny threshold based on the Otsu binarization of the image gradient is an effective approach for an image sequence.

\begin{figure}[h]
\centering
\includegraphics[width=0.9\linewidth]{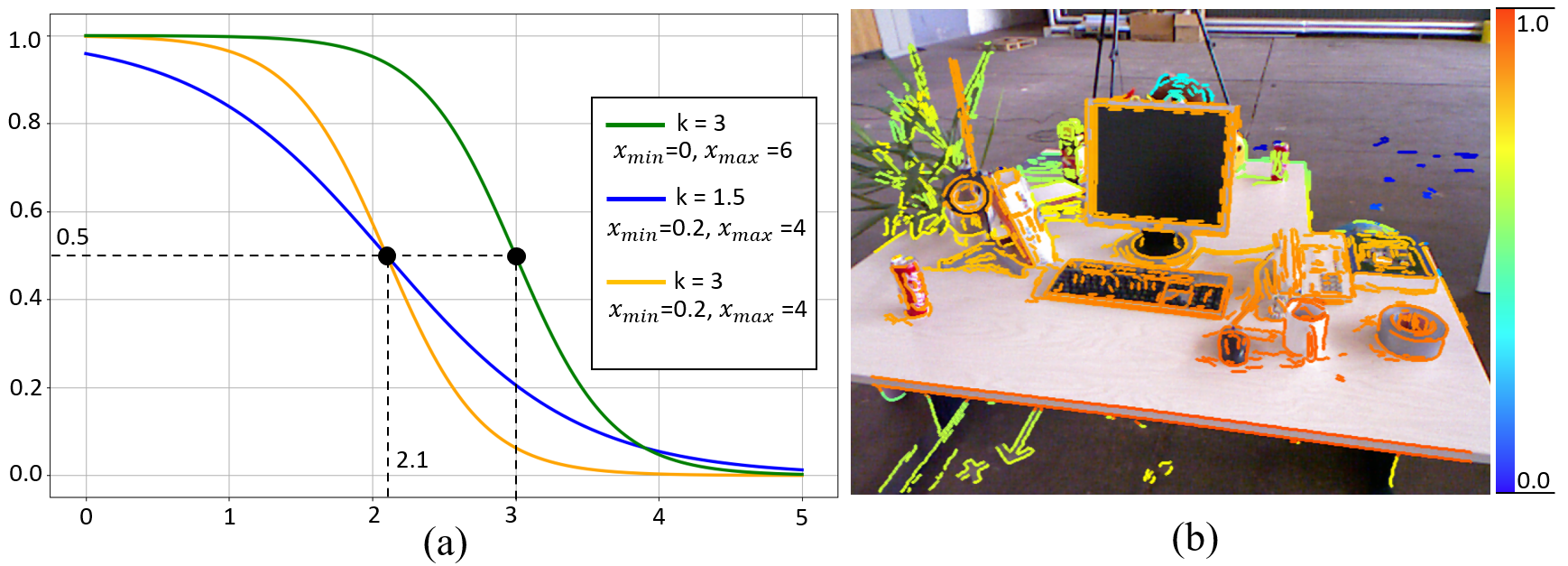}
\caption{ Illustration of depth-based weighting scheme. (a) Function curves of $w_s$ under different parameter configurations. (b) Pixel-wise weights of a real-world image for pose optimization.}
\label{ws}
\end{figure}

Additionally, we hereby introduce the adaptive weight for robust optimization mentioned in Eq. \ref{coarseWeight}. Due to the noise in the depth measurements of the real-world RGB-D cameras, points distant from the camera tend to have higher levels of noise in depth measurements. As a result, when performing least squares optimization in coarse tracking, fine tracking, and bundle adjustment, it is essential to weigh the residuals based on the distance of the pixels from the camera such that farther points correspond to lower weights. To determine an appropriate weighting function, we compared several different weighting schemes and chose the following formula:
\begin{equation}
\label{sensor_weight}
w_s(d) = \frac{1}{1 + e^{k(d-(d_{max}+d_{min})/2)}},
\end{equation}
where $d$ is the distance of the pixel from the camera, $d_{min}, d_{max}$ are the minimum and maximum effective measurement depths of the camera, respectively, and $k$ is a constant related to the camera's depth accuracy. If a camera has greater depth noise, $k$ should be correspondingly larger. Curves of $w_s(d)$ under different parameter configurations are shown in Fig. \ref{ws} (a), and the weight of the edges during optimization is shown in Fig. \ref{ws} (b). We found that applying this weighting function effectively improves the quality of pose estimation. Moreover, the configuration can be easily adjusted according to the tech specs of the sensor. For the public datasets: ICL-NUIM \cite{icl}, TUM RGB-D \cite{tum}, and ETH-3D \cite{eth_3d}, we set the parameters $(k, d_{min}, d_{max})$ as (0.5, 0.1, 6), (0.75, 0.2, 4.5), and (0.5, 0.15, 5) respectively according to each dataset's depth image quality. These parameter configurations remained consistent throughout all evaluations within the same dataset.

\subsection{Real-world and Synthetic Experiments}

In this subsection, we compare the performance of our method with the state-of-the-art algorithms across different scenes from several publicly available benchmarks, including the real-world TUM RGB-D benchmark \cite{tum}, the ETH-3D SLAM benchmark \cite{eth_3d}, and the synthetic ICL-NUIM dataset \cite{icl}. Additionally, we tested our approach on our own dataset collected in conventional real-world indoor and outdoor environments with sparse textures and common natural disturbances.

To ensure a fair and comprehensive comparison, we compared our method against various state-of-the-art approaches, each employing distinct feature representation approaches and pose optimization techniques. These include ORB-SLAM3 \cite{orb-slam3} (RGBD version), which relies on sparse feature points, edge feature-based methods like CannyVO \cite{cannyvo}, and Re-SLAM \cite{re_slam}. These RGB-D visual odometry algorithms utilize edge features through distance fields,  Manhattan-SLAM \cite{manhattanslam}, which leverages ideal straight line and plane features for pose estimation in indoor scenes. Furthermore, we also conduct comparative evaluations against recently emerging learning-based vSLAM pipelines, including DROID-SLAM \cite{droid-slam} that estimates camera pose through a neural solver, Photo-SLAM \cite{photo-slam} that combines 3D Gaussian Splatting with geometric feature points, and MAST3R-SLAM \cite{mast3r-slam} that leverages learned reconstruction priors. Among them, DROID-SLAM and Photo-SLAM support RGB-D mode, while MAST3R-SLAM operates solely in RGB mode but achieves realistic depth perception by leveraging a powerful depth prior model. Additionally, in order to evaluate the contribution of each module in both our method and the baseline methods to the pose estimation results, we conducted experiments with modular comparisons. For ORB-SLAM3, we tested its performance in two modes: tracking-only mode and local mapping mode, denoted as \textit{ORB-SLAM3-TR} and \textit{ORB-SLAM3-LM}, respectively. To make ORB-SLAM3 perform tracking-only mode, we set the variable \texttt{mbOnlyTracking} to be \texttt{true} in the open-source code. To run it with local mapping, we disabled the loop closure thread while keeping other settings unchanged. Since our method is purely a visual odometry (VO) approach without loop closure detection or global optimization, we maintain consistent loop-free configurations when evaluating other methods, for DROID-SLAM \cite{droid-slam}, we execute its front end including the tracking and local BA modules, for Photo-SLAM \cite{photo-slam}, since its pose estimation relies on ORB-SLAM3 \cite{orb-slam3}, we disable the loop closure module in the ORB-SLAM3 component. And for MAST3R-SLAM \cite{mast3r-slam}, we retained its original configuration as it doesn't have explicit front-end/back-end separation. Since MAST3R-SLAM \cite{mast3r-slam} gives trajectories lacking metric scale, we evaluated its performance using scale-aligned trajectories in our assessment. Additionally, since MAST3R-SLAM\cite{mast3r-slam} provides multiple models of varying sizes, we employ the medium-sized model for experiments to balance quality and efficiency. For our method, we tested its performance in three configurations: coarse tracking only, full coarse-to-fine tracking only, and complete odometry with edge fusion and bundle adjustment, denoted as \textit{ROEVO-CT}, \textit{ROEVO-TR}, and \textit{ROEVO-TR+BA}, respectively. Using the trajectory errors obtained under these three different configurations, we can conduct an ablation study to evaluate the contribution of each proposed module to system accuracy.

Since our VO system does not include loop closure detection to correct trajectory drift over time, we evaluated the performance using both the \textit{Relative Pose Error} (RPE) and \textit{Absolute Trajectory Error} (ATE). The RPE primarily reflects the odometry drifts within a local range, it can be further subdivided into rotational RPE (deg/s), indicating rotational drift and translational RPE (cm/s) which represents positional deviation. The ATE mainly assesses the global consistency of the trajectories, given the relatively short length of the RGB-D data sequences.

\subsubsection{ICL-NUIM Dataset}
The ICL-NUIM dataset is a synthetic RGB-D benchmark providing high-quality rendered images and depth measurements for visual localization and scene reconstruction. The scenes in the datasets are set in a simplistic living room environment, where sparse texture and rapidly changing viewpoints pose challenges for visual odometry. Some sequences in the ICL-NUIM dataset contain viewpoints that only capture bare walls, making it impossible to extract meaningful edge features for pose estimation. Therefore, we truncated these problematic sequence segments and excluded the frames where pose estimation is not feasible. If a sequence is truncated into multiple segments, the final trajectory error is computed as the weighted mean of these segments, with weights proportional to the frame number in each segment. For fairness, all baseline methods used the same truncated sequence segments in the experiments.

\begin{figure}[h]
\centering
\includegraphics[width=0.9\linewidth]{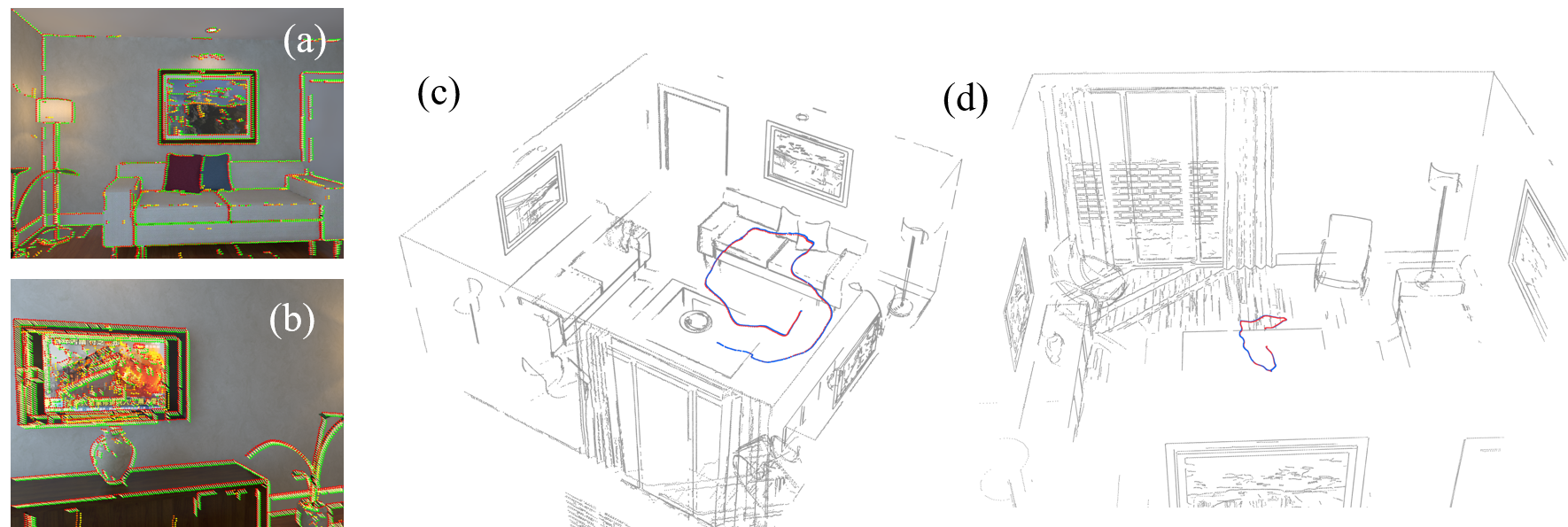}
\caption{Qualitative results of the proposed ROEVO on the synthetic ICL-NUIM dataset. (a) and (b) show the tracking process with the keyframe's edges in green and their corresponding associated current frame's edges in red. (c) and (d) are the estimated trajectories with semi-dense reconstruction from sequences \textit{lr\_kt2} and \textit{lr\_kt1}, respectively.}
\label{icl_traj}
\end{figure}

Table \ref{ate_icl} presents the Root Mean Square Error (RMSE) of absolute trajectory error for different algorithms on the ICL-NUIM datasets, and Table \ref{rpe_icl_full} presents the RMSE of RPE. We use ``*" superscript to indicate that MAST3R-SLAM is running in RGB mode, and the subsequent tables will follow this notation. The bold result with green background represents the best-performing method among all, the underlined result with green background represents the second-best-performing method among all, and the result with yellow background represents the third. Due to the extreme lack of texture in this dataset, we had to significantly lower the FAST corner threshold of ORB-SLAM3 to ensure it functioned properly. Fig. \ref{icl_traj} shows the trajectories estimated by our method and the reconstructed maps. In this dataset, leveraging structural information proved to be highly effective in reducing trajectory errors. For instance, Manhattan-SLAM \cite{manhattanslam}, which utilizes line and plane features, performed better than ORB-SLAM3 relying solely on texture features. By extracting structured edges and incorporating them into fine tracking, our method achieved substantial improvements in tracking accuracy. Furthermore, our organized edge feature-based BA framework contributed significantly to the overall accuracy by fully utilizing and optimizing the spatial structure of edges. As a result, our full method outperformed both ORB-SLAM3 and Manhattan-SLAM in the absolute trajectory errors.

It is worth noting that our method is slightly inferior to Manhattan-SLAM on the \textit{lr\_kt3} sequence. This is attributed to the presence of numerous specular reflections in \textit{lr\_kt3}, which significantly affect methods relying on texture or edge information. Nevertheless, our method effectively mitigates the odometry drift by specifying and rejecting outlier edge observations during bundle adjustment, leading to accurate corrections of both multi-frame poses and spatial edges in the local map over time. Our approach demonstrates superior performance on this dataset compared to all baseline methods. This advantage stems from two key properties of organized edges: first, organized edges remain reliably detectable even in structure-sparse environments where conventional features fail, and second, our framework enables explicit edge feature association, hence achieving higher accuracy compared to other edge-based approaches such as CannyVO \cite{cannyvo} and RE-SLAM \cite{re_slam} that do not distinguish edges.

\begin{table}[t]
\caption{Absolute Trajectory RMSE (CM) on ICL-NUIM datasets\label{ate_icl}}
\centering
\setlength\tabcolsep{6pt}
\begin{tabular}{c|cccc|c}
\toprule
Method/Seq. & lr\_kt0 & lr\_kt1 & lr\_kt2 & lr\_kt3 & average \\
  \cmidrule{1-6}
  ORB-SLAM3-TR & 3.486 &  6.100 & 9.899 & 13.65 & 8.282\\
  ORB-SLAM3-LM\cite{orb-slam3} & \third{0.552} &  2.609 & \third{1.558} & 1.023 & 1.435 \\
  CannyVO\cite{cannyvo} & 3.500 &  0.900 & 3.100 & 8.000 & 3.875\\
 ManhattanSLAM\cite{manhattanslam}& 0.609 &  0.852 & 1.786 & \first{0.332} & \second{0.894}\\
 RE-SLAM\cite{re_slam} & 0.873 &  1.520 & 2.144 & 3.197 & 1.934\\
 DROID-SLAM\cite{droid-slam} & \second{0.521} &  \third{0.766} & 0.571 & 11.16 & 3.254\\
 PHOTO-SLAM\cite{photo-slam} & 0.701 &  15.05 & 1.702 & \third{0.665} & 4.529\\
 MAST3R-SLAM\textsuperscript{*} \cite{mast3r-slam} & 0.982 &  0.876 & \second{1.464} & 1.344 & 1.166\\
 \cmidrule{1-6}
 ROEVO-CT & 0.857 &  1.262 & 3.444 & 1.378 & 1.735\\
 ROEVO-TR & 0.624 &  \second{0.491} & 1.721 & 1.264 & \third{1.025}\\
 ROEVO-TR+BA & \first{0.347} &  \first{0.399} & \first{1.125} & \second{0.659} & \first{0.633}\\
\bottomrule
\end{tabular}
\end{table}

\begin{table*}[!b]
\caption{Relative Trajectory RMSE (R: DEG/S, t: CM/S) on ICL-NUIM\cite{icl} Datasets}
\label{rpe_icl_full}
\centering
\setlength\tabcolsep{3pt}
\begin{tabular}{c|cc|cc|cc|cc|cc}
\toprule
  & \multicolumn{2}{c}{lr\_kt0} & \multicolumn{2}{c}{lr\_kt1} & \multicolumn{2}{c}{lr\_kt2} & \multicolumn{2}{c}{lr\_kt3} & \multicolumn{2}{c}{average} \\
 
\cmidrule{1-11}
Seq. & RMSE (R) & RMSE (t)  & RMSE (R) & RMSE (t) & RMSE (R) & RMSE (t) & RMSE (R) & RMES (t) & RMSE (R) & RMSE (t)\\
\cmidrule{1-11}

ORB-SLAM3-TR &0.532 &  1.931 & 0.358 & 1.820 & 0.705 & 2.941 & 0.281 & 0.989 &  0.469 & 1.920\\
  ORB-SLAM3-LM\cite{orb-slam3} & \third{0.111} &  0.566 & 0.567 & 2.491 & 0.164 & 0.863 & \third{0.104} & 0.459 &  0.236 & 1.094 \\
  CannyVO\cite{cannyvo} & 0.674 &  1.400 & 0.208 & 0.900 & 0.269 & 1.100 & 0.152 & 0.700 &  0.325 & 1.025\\
 ManhattanSLAM\cite{manhattanslam}& 0.116 &  0.535 & 0.154 & 0.685 & 0.179 & 0.931 & \second{0.086} & \second{0.325} &  \third{0.134} & 0.619 \\
 RE-SLAM\cite{re_slam} & 0.211 &  0.997 & 0.197 & 0.804 & 0.243 & 1.029 & 2.280 & 3.549 &  0.732 & 1.594 \\
 DROID-SLAM\cite{droid-slam} & 0.044 &  \second{0.268} & \third{0.052} & \third{0.274} & \second{0.062} & \second{0.356} & 0.578 & 3.611 &  0.184 & 1.127 \\
 PHOTO-SLAM\cite{photo-slam} & 0.234 &  0.833 & 1.669 & 7.267 & 0.178 & 0.913 & 0.158 & 0.752 &  0.559 & 2.441 \\
 MAST3R-SLAM\textsuperscript{*}\cite{mast3r-slam} & 0.166 &  0.443 & 0.099 & 0.329 & 0.233 & 0.891 & 0.254 & \third{0.361} &  0.188 & \third{0.506} \\
 \cmidrule{1-11}
 ROEVO-CT & 0.265 &  0.736 & 0.102 & \second{0.194} & 0.173 & 1.096 & 0.106 & 0.555 &  0.145 & 0.645\\
 ROEVO-TR & \second{0.106} &  \third{0.361} & \first{0.037} & 0.430 & \third{0.107} & \third{0.564} & 0.130 & 0.526 &  \second{0.111} & \second{0.470} \\
 ROEVO-TR+BA & \first{0.042} &  \first{0.214} & \second{0.043} & \first{0.185} & \first{0.057} & \first{0.336} & \first{0.061} & \first{0.284} &  \first{0.051} & \first{0.254} \\
\bottomrule
\end{tabular}
\end{table*}

\subsubsection{TUM Dataset} 

\begin{figure}[h]
\centering
\includegraphics[width=0.9\linewidth]{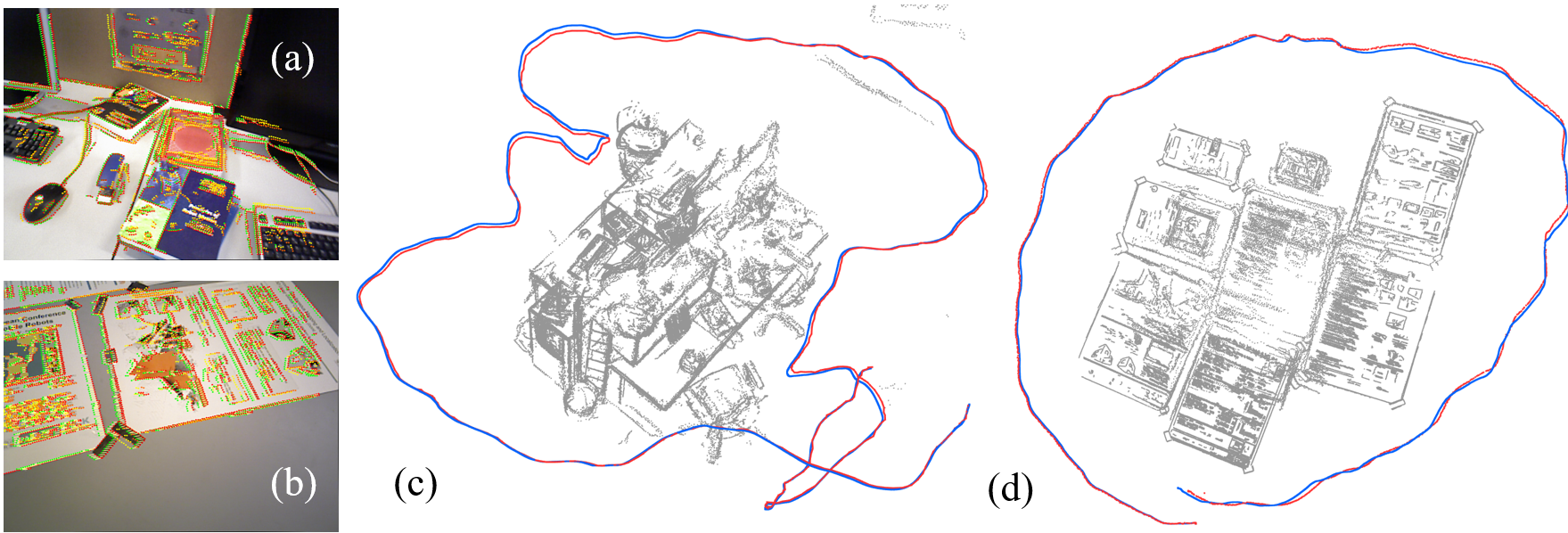}
\caption{Performance from the synthetic TUM-RGBD dataset. (a) and (b) show the tracking process with the keyframe's edges in green and their corresponding associated current frame's edges in red. (c) and (d) are the trajectory results with semi-dense reconstruction, where (c)\textit{fr3\_long\_office} (d)\textit{fr3\_no\_structure\_texture\_near}}
\label{tum_traj}
\end{figure}

We also conducted experiments on the TUM RGB-D dataset \cite{tum}, a widely used benchmark for evaluating RGB-D VO. We primarily tested the commonly used handheld SLAM sequences and typical structure-texture sequences. For evaluation, we calculated both the RPE and ATE for the resulting trajectories. The RPE results are presented in Tables \ref{rpe_r_tum} and \ref{rpe_t_tum}, and the ATE results are shown in Table \ref{ate_tum}. The trajectories and reconstruction results for these sequences are shown in Fig. \ref{tum_traj}.

From the evaluation results, we can see that our method significantly improves the translational results of coarse tracking through fine tracking. This enhancement is due to the utilization of the spatial structure of edges in fine tracking, while coarse tracking relies solely on image textures. This highlights the importance of edge spatial structure in pose estimation. Additionally, we find that BA has a notable improvement in optimizing pose rotation. BA places a strong emphasis on the spatial consistency of multi-frame edges, which improves rotation optimization. From Tables \ref{ate_tum}, \ref{rpe_t_tum}, and \ref{rpe_r_tum},  we can see that our method demonstrates superior performance compared to other baselines on TUM RGB-D's \textit{structure-texture} sequences. However, we also observed that for handheld SLAM sequences such as \textit{fr1\_desk}, \textit{fr2\_desk}, and \textit{fr3\_office}, ORB-SLAM3 \cite{orb-slam3}, RE-SLAM \cite{re_slam}, and Photo-SLAM \cite{photo-slam} exhibit smaller errors in the ATE. This occurs because the unique trajectory patterns in these sequences as they contain multiple viewpoints at different segments of the trajectory, observing the same locations. While SLAM methods that preserve all keyframes for local mapping exploit these long-term constraints for better ATE, our sliding-window approach maintains competitive RPE by preserving local accuracy through the co-visibility graph of organized edges. Nevertheless, our approach still outperforms other geometry-based methods like CannyVO and Manhattan-SLAM on these sequences. In this dataset, learning-based methods such as MAST3R-SLAM \cite{mast3r-slam} and DROID-SLAM \cite{droid-slam} demonstrate poor performances. This is primarily attributed to two inherent dataset characteristics like the lack of strict temporal synchronization between depth and RGB channels, and the presence of significant noise in depth measurements. These acquisition conditions prove particularly challenging for learning-based approaches to process effectively. Our method consistently outperforms learning-based approaches across all evaluated sequences, further demonstrating its robustness under challenging, suboptimal data. In addition, since our method prioritizes ensuring the consistency of structure and texture during optimization, we achieve superior performance on the structure-texture sequences.

\begin{table*}[h]
\caption{Absolute Trajectory RMSE (cm) on TUM-RGBD\cite{tum} Datasets\label{ate_tum}}
\centering
\setlength\tabcolsep{6pt}
\begin{tabular}{c|cccccccc|c}
\toprule
Method/Seq. & fr1\_desk & fr2\_xyz & fr2\_desk & fr3\_office & fr3\_nst\_tex\_n & fr3\_nst\_tex\_f & fr3\_str\_tex\_f & fr3\_str\_tex\_n & average \\
  \cmidrule{1-10}
  ORB-SLAM3-TR & 6.879 & 0.831 & 38.93 & 40.23 & 11.13 & 14.45 & 3.000 & 4.541 & 14.987\\
  ORB-SLAM3-LM\cite{orb-slam3} & \second{1.766} & \second{0.314} & \first{0.921} & \second{1.157} & \second{2.735} & 3.474 & \third{1.333} & 1.552 & \second{1.656}\\
  ManhattanSLAM\cite{manhattanslam} & 7.300 & 1.112 & 2.945 & 9.688 & 5.234 & 3.991 & 2.477 & \third{1.411} & 4.269\\
  CannyVO\cite{cannyvo}&  4.400 & 0.800 & 3.700 & 8.500 & 9.00 & \third{2.600} & 1.200 & 2.500 & 4.087\\
 RE-SLAM\cite{re_slam} & \third{3.152} &  \third{0.527} & \third{1.782} & 3.798 & 27.74 & 36.68 & 1.649 & 1.314 & 9.580\\
 MAST3R-SLAM\textsuperscript{*}\cite{mast3r-slam}& 9.256 & 3.201 & 8.503 & 23.06 & 46.14 & 56.71 & 51.84 & 20.71& 27.43\\
 DROID-SLAM\cite{droid-slam} & 5.354 & 6.929 & 27.63 &  38.28 & 8.419 & 5.678 & 5.227 & 9.023 & 13.32\\
 PHOTO-SLAM\cite{photo-slam} & \first{1.613} & \first{0.311} & \second{1.045} & \first{1.112} & \third{2.946} & 2.937 & 1.866 & \second{1.391} & \first{1.653}\\
 \cmidrule{1-10}
 ROEVO-CT & 6.576 & 1.141 & 3.211 & 5.956 & 4.194 & 3.503 & 3.091 & 3.957 & 3.952\\
 ROEVO-TR & 6.041 & 0.772 & 2.231 & 4.247 & 4.167 & \second{2.467} & \second{1.288} & 1.621 & 2.854\\
 ROEVO-TR+BA & 5.436 & 0.819 & 2.756 & \third{2.331} & \first{2.718} & \first{2.192} & \first{0.882} & \first{1.296} & \third{2.303}\\
\bottomrule
\end{tabular}
\end{table*}

\begin{table*}[h]
\caption{Relative Translational RMSE (cm/s) on TUM RGB-D datasets\label{rpe_t_tum}}
\centering
\setlength\tabcolsep{6pt}
\begin{tabular}{c|cccccccc|c}
\toprule
Method/Seq. & fr1\_desk & fr2\_xyz & fr2\_desk & fr3\_office & fr3\_nst\_tex\_n & fr3\_nst\_tex\_f & fr3\_str\_tex\_f & fr3\_str\_tex\_n & average \\
  \cmidrule{1-10}
  ORB-SLAM3-TR & 5.133 & 0.521 & 3.122 & 2.545 & 4.412 & 12.11 & 1.866 & 1.732 & 3.931\\
  ORB-SLAM3-LM\cite{orb-slam3} & \third{2.433} & 0.312 & \third{0.631} & \second{0.918} & 2.313 & 4.871 & 1.619 & 1.477 & 1.822 \\
  ManhattanSLAM\cite{manhattanslam} & 6.718 & 5.388 & 1.143 & 4.627 & 1.531 & \second{3.129} & 1.382 & 1.299 & 3.152\\
  CannyVO\cite{cannyvo} & 3.100 & 0.300 & \third{0.800} & 1.000 & 2.900 & \third{3.500} & \second{1.200} & \second{1.000} & 1.725\\
 RE-SLAM\cite{re_slam} & 3.289 & 0.437 & 1.026 & 1.075 & 7.573 & 38.69 & 1.668 & 1.122 & 6.860\\
 MAST3R-SLAM\textsuperscript{*}\cite{mast3r-slam} & 4.913 & 0.385 & 0.890 & 2.563 & 3.059 & 24.95 & 1.622 & 1.349 & 4.966\\
 DROID-SLAM\cite{droid-slam}  &3.794 & 0.295 & 1.767 & 1.394 & 3.371 & 21.35 & 1.965 & 1.949 & 4.485\\
 PHOTO-SLAM\cite{photo-slam} & \first{2.086} & \second{0.242} & \second{0.661} & \first{0.831} & \first{1.312} & 3.789 & \third{1.223} & \third{1.011} & \first{1.394}\\
 \cmidrule{1-10}
 ROEVO-CT & 2.965 & 0.465 & 0.968 & 1.047 & 1.185 & 3.817 & 1.689 & 1.927 & 1.757\\
 ROEVO-TR & \second{2.147} & \third{0.288} & 0.845 & 1.035 & \second{1.377} & 3.553 & 1.322 & 1.142 & \third{1.463}\\
 ROEVO-TR+BA & 2.605 & \first{0.241} & 0.821 & \third{0.965} & \third{1.380} & \first{3.051} & \first{1.176} & \first{0.977} & \second{1.402}\\
\bottomrule
\end{tabular}
\end{table*}

\begin{table*}[h]
\caption{Relative Rotational RMSE (deg/s) on TUM RGB-D datasets\label{rpe_r_tum}}
\centering
\setlength\tabcolsep{6pt}
\begin{tabular}{c|cccccccc|c}
\toprule
Method/Seq. & fr1\_desk & fr2\_xyz & fr2\_desk & fr3\_office & fr3\_nst\_tex\_n & fr3\_nst\_tex\_f & fr3\_str\_tex\_f & fr3\_str\_tex\_n & average \\
  \cmidrule{1-10}
  ORB-SLAM3-TR & 2.389 & 0.323 & 1.112 & 0.926 & 1.600 & 2.316 & 0.618 & 0.890 & 1.272\\
  ORB-SLAM3-LM\cite{orb-slam3} & \third{1.653} & \first{0.195} & \first{0.406} & 0.487 & 1.097 & 1.011 & 0.575 & 0.678 & 0.762\\
  ManhattanSLAM\cite{manhattanslam} & 3.152 & 0.316 & 0.522 & 1.958 & 0.873 & \second{0.806} & \second{0.443} & 0.688 & 1.095\\
  CannyVO\cite{cannyvo} & 1.923 & \second{0.307} & \third{0.458} & 0.503 & 1.440 & 0.892 & 0.459 & 0.593 & 0.822\\
 RE-SLAM\cite{re_slam} & 2.215 & 0.337 & 0.567 & 0.616 & 3.246 & 1.229 & 0.637 & \third{0.631} & 1.185\\
 MAST3R-SLAM\textsuperscript{*}\cite{mast3r-slam} & \second{1.409} & 0.374 & 0.488 & 0.456 & 0.866 & 1.745 & 0.633 & 0.709 & 0.835\\
 DROID-SLAM\cite{droid-slam} & 3.111 & 0.491 & 1.257 & 0.781 & 0.819 & 2.144 & 0.875 & 0.793 & 1.283\\
 PHOTO-SLAM\cite{photo-slam} & \first{1.343} & 0.231 & \second{0.442} & \third{0.485} & \third{0.766} & 0.963 & 0.502 & \second{0.606} & \first{0.667}\\
 \cmidrule{1-10}
 ROEVO-CT & 1.957 & 0.328 & 0.525 & 0.504 & 0.771 & 0.902 & 0.565 & 0.916 & 0.808\\
 ROEVO-TR & 1.922 & 0.310 & 0.482 & \second{0.484} & \second{0.757} & \third{0.821} & \third{0.481} & 0.640 & \third{0.737}\\
 ROEVO-TR+BA & 1.919 & \third{0.308} & 0.462 & \first{0.472} & \first{0.744} & \first{0.747} & \first{0.419} & \first{0.552} & \second{0.703}\\
\bottomrule
\end{tabular}
\end{table*}

\subsubsection{ETH-3D Dataset}

We also conducted experiments on the ETH-3D dataset\cite{eth_3d}, which contains numerous real-world sequences collected by RGB-D cameras. Unlike the rolling shutter and the asynchronous RGB-depth channel issues presented in some datasets, ETH-3D sequences benefit from better channel alignment and RGB images captured by a global shutter camera, ensuring more stable pose estimation results. ETH-3D offers particularly challenging sequences for feature extraction, as well as sequences with sparse textures, making it highly suitable for evaluating the accuracy and robustness of VO systems. We evaluate both APEs and RPEs of the proposed method and other baselines in Table \ref{rpe_r_eth}, \ref{rpe_t_eth}, and Table \ref{ate_eth}, respectively.

\begin{figure}[h]
\centering
\includegraphics[width=0.9\linewidth]{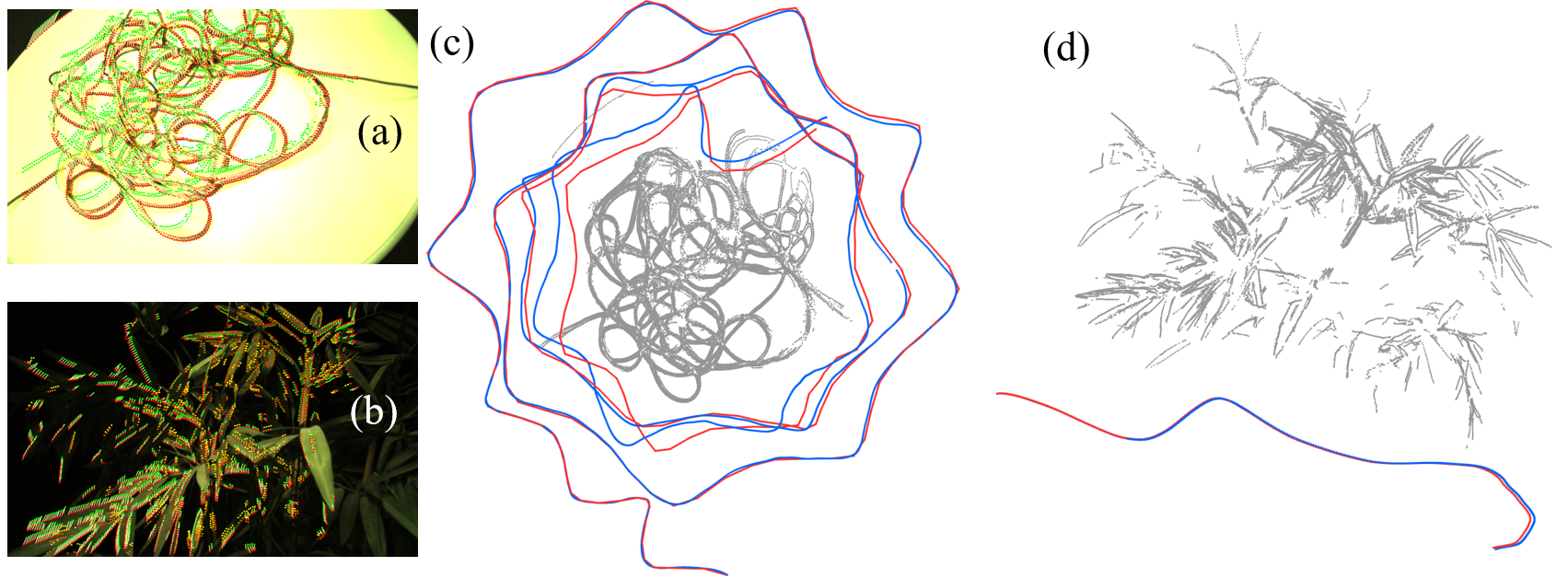}
\caption{Qualitative results of the proposed ROEVO on the ETH-3D dataset. (a) and (b) show the tracking process with the keyframe's edges in green and their corresponding associated current frame's edges in red. (c) and (d) are the estimated trajectories with semi-dense reconstruction from sequences \textit{cables\_1} and \textit{plant\_2}, respectively. (c) and (d) are the estimated trajectories with semi-dense reconstruction from sequences \textit{lr\_kt2} and \textit{lr\_kt1}, respectively.}
\label{eth_traj}
\end{figure}

As shown in the tables, our method demonstrates robust performance in sequences with sparse textures and significant occlusion, such as \textit{plant\_2}, \textit{sofa\_3}, and \textit{plant\_scene}. The trajectories and reconstruction results for these sequences are demonstrated in Fig. \ref{eth_traj}. In environments with extremely sparse textures, our algorithm maintains high robustness and completes all sequences in the ETH-3D dataset, whereas geometry-based methods like Manhattan-SLAM and ORB-SLAM3 (tracking only) and learning-assisted methods like Photo-SLAM fail to maintain robustness under such extreme conditions. Our method also consistently outperforms other methods in terms of ATE error. Although ORB-SLAM3 performs well in texture-rich environments like \textit{repetitive} and \textit{planar\_3}, it cannot maintain the same level of performance in texture-sparse conditions. We also notice that in some sequences, such as \textit{cables\_1}, our method achieves smaller RPE than ORB-SLAM3 but does not show as large an improvement in ATE compared to other methods. This is because the camera moves within a limited small range in \textit{cables\_1} sequence, even though we disabled loop closure in ORB-SLAM3, it still benefits from local map associations in such conditions, which effectively function like global optimization, helping to produce a trajectory with smaller global errors. In the ETH-3D dataset, learning-based DROID-SLAM achieves the highest accuracy. This is primarily because the data distribution in ETH-3D closely resembles that of DROID-SLAM's training data. However, on benchmarks like ICL-NUIM and TUM RGB-D, DROID-SLAM does not exhibit similar performance. This reflects a broader limitation of learning-based approaches that their accuracy tends to be inconsistent across different benchmarks, unlike geometry-based methods, which provide more consistent results.

\begin{table*}[t]
\caption{Absolute Trajectory RMSE (cm) on ETH-3D Datasets\label{ate_eth}}
\centering
\setlength\tabcolsep{6pt}
\begin{tabular}{c|cccccccc|c}
\toprule
Method/Seq. & cables\_1 & plant\_2 & repetitive & planar\_3 & sofa\_3 & plant\_scene\_1 & plant\_scene\_2 & plant\_scene\_3 & average \\
  \cmidrule{1-10}
  ORB-SLAM3-TR & 5.386 & 1.276 & 4.386 & 2.961 & $\times$ & $\times$ & 22.79 & 45.45 & 13.71\\
  ORB-SLAM3-LM\cite{orb-slam3} & \third{0.807} &  \second{0.170} & \first{0.285} & \first{0.669} & 27.33 & \second{1.278} & 2.784 & 6.399 & 4.965\\
  ManhattanSLAM\cite{manhattanslam}& 0.991 &  0.360 & \second{0.579} & \second{0.877} & $\times$ & 4.991 & 3.551 & 7.776 & 2.732\\
 RE-SLAM\cite{re_slam} & \first{0.641} &  0.245 &  0.765 & 11.577  & $\times$ & 9.059 & \third{1.531} & 37.58 & 8.771\\
 DROID-SLAM\cite{droid-slam} & 1.872 & 0.478 & \third{0.715} & \third{1.147} & \first{0.788} & \third{1.486} & 1.985 & \first{0.704} & \first{1.147}\\
 PHOTO-SLAM\cite{photo-slam} & \second{0.701} &  \third{0.194} & $\times$ & 1.416 & \second{1.095} & 20.95 & \second{1.325} & 3.511 & 4.142\\
 MAST3R-SLAM\textsuperscript{*}\cite{mast3r-slam} & 3.302 &  2.934 & 12.98 & 16.70 & 11.64 & 16.61 & 9.31 & 13.60 & 10.885\\
 \cmidrule{1-10}
 ROEVO-CT & 2.438 &  0.475 & 4.172 & 5.160 & 7.925 & 5.471 & 3.614 & 2.679 & 3.992\\
 ROEVO-TR & 1.823 &  0.213 & 2.508 & 2.712 & 6.149 & 3.163 & 1.903 & \third{2.35} & \third{2.603}\\
 ROEVO-TR+BA & 1.322 &  \first{0.166} & 1.335 & 2.351 & \third{4.388} & \first{0.987} & \first{1.311} & \second{1.273} & \second{1.642}\\
\bottomrule
\end{tabular}
\end{table*}

\begin{table*}[t]
\caption{Relative Rotational RMSE (deg/s) on ETH-3D datasets\label{rpe_r_eth}}
\centering
\setlength\tabcolsep{6pt}
\begin{tabular}{c|cccccccc|c}
\toprule
Method/Seq. & cables\_1 & plant\_2 & repetitive & planar\_3 & sofa\_3 & plant\_scene\_1 & plant\_scene\_2 & plant\_scene\_3 & average \\
  \cmidrule{1-10}
  ORB-SLAM3-TR & 1.520 & 0.880 & 2.042 & 1.001 & $\times$ & $\times$ & 3.347 & 7.034 & 2.637 \\
  ORB-SLAM3-LM\cite{orb-slam3} & 0.969 &  0.314 & \second{0.329} & 1.084 & 4.331 & 0.517 & 0.397 & 0.769 & 1.089 \\
  ManhattanSLAM\cite{manhattanslam}& 0.867 &  0.398 & \third{0.478} & 0.589 & $\times$ & 0.717 & 0.472 & 1.932 & 0.779\\
 RE-SLAM\cite{re_slam} & \second{0.504} &  \third{0.222} &  0.689 & 1.492  & $\times$ & 0.866 & 0.374 & 2.552 & 0.957\\
 DROID-SLAM\cite{droid-slam} & \first{0.438} & 0.261 & \first{0.189} & \first{0.233} & \first{0.251} & \first{0.206} & \first{0.214} & \first{0.199} & \first{0.249} \\
 PHOTO-SLAM\cite{photo-slam} & 0.709 & 0.306 & $\times$ & 0.729 & \second{0.683} & \third{0.367} & \third{0.346} & 0.661 & \third{0.543}\\
 MAST3R-SLAM\textsuperscript{*}\cite{mast3r-slam} & \third{0.679} & 3.055 & 2.330 & 2.766 & 1.988 & 1.496 & 0.781 & 3.839 & 2.117\\
 \cmidrule{1-10}
 ROEVO-CT & 0.822 & 0.368 & 1.714 & 0.960 & 2.017 & 0.835 & 0.393 & 0.991 & 1.013\\
 ROEVO-TR & 0.753 & \second{0.217} & 1.175 & \third{0.524} & 1.504 & 0.492 & 0.384 & \third{0.458} & 0.688\\
 ROEVO-TR+BA & 0.721 & \first{0.213} & 0.894 & \second{0.418} & \third{1.082} & \second{0.317} & \second{0.255} & \second{0.419} & \second{0.539}\\
\bottomrule
\end{tabular}
\end{table*}

\begin{table*}[t]
\caption{Relative Translational RMSE (cm/s) on ETH-3D datasets\label{rpe_t_eth}}
\centering
\setlength\tabcolsep{6pt}
\begin{tabular}{c|cccccccc|c}
\toprule
Method/Seq. & cables\_1 & plant\_2 & repetitive & planar\_3 & sofa\_3 & plant\_scene\_1 & plant\_scene\_2 & plant\_scene\_3 & average \\
  \cmidrule{1-10}
  ORB-SLAM3-TR & 12.45 & 11.91 & 2.063 & 1.571 & $\times$ & $\times$ & 13.71 & 33.28 & 12.49\\
  ORB-SLAM3-LM\cite{orb-slam3} & 0.878 &  0.325 & \second{0.374} & 1.679 & 12.51 & 2.972 & 1.548 & 2.633 & 2.865\\
  ManhattanSLAM\cite{manhattanslam} & 0.845 & 0.433 & \third{0.441} & \second{0.955} & $\times$ & 2.626 & 2.173 & 8.177 & 2.236\\
 RE-SLAM\cite{re_slam} & \second{0.461} &  \third{0.291} &  0.738 & 3.281  & $\times$ & 2.916 & 1.629 & 12.45 & 3.109\\
 DROID-SLAM\cite{droid-slam} & \first{0.359} & 0.432 & \first{0.264} & \first{0.306} & \first{0.548} & \first{0.608} & \first{0.637} & \first{0.476} & \first{0.454}\\
 PHOTO-SLAM\cite{photo-slam} & 0.706 & 0.323 & $\times$ & 1.132 & \second{1.632} & \third{1.262} & \third{1.218} & 2.934 & \third{1.315}\\
 MAST3R-SLAM\textsuperscript{*}\cite{mast3r-slam} & 4.500 & 21.59 & 10.41 & 48.68 & 10.04 & 16.54 & 5.17 & 22.01 & 17.37\\
 \cmidrule{1-10}
 ROEVO-CT & 0.642 & 0.509 & 1.613 & 2.067 & 5.264 & 2.812 & 1.438 & 3.538 & 2.235\\
 ROEVO-TR & 0.593 & \second{0.239} & 1.102 & 1.589 & 3.785 & 1.799 & 1.639 & \third{1.822} & 1.571\\
 ROEVO-TR+BA & \third{0.576} & \first{0.211} & 0.966 & \third{1.131} & \third{2.742} & \second{0.898} & \second{0.866} & \second{1.164} & \second{1.069}\\
\bottomrule
\end{tabular}
\end{table*}

\subsubsection{Chair Dataset}
\begin{figure}[h]
\centering
\includegraphics[width=0.9\linewidth]{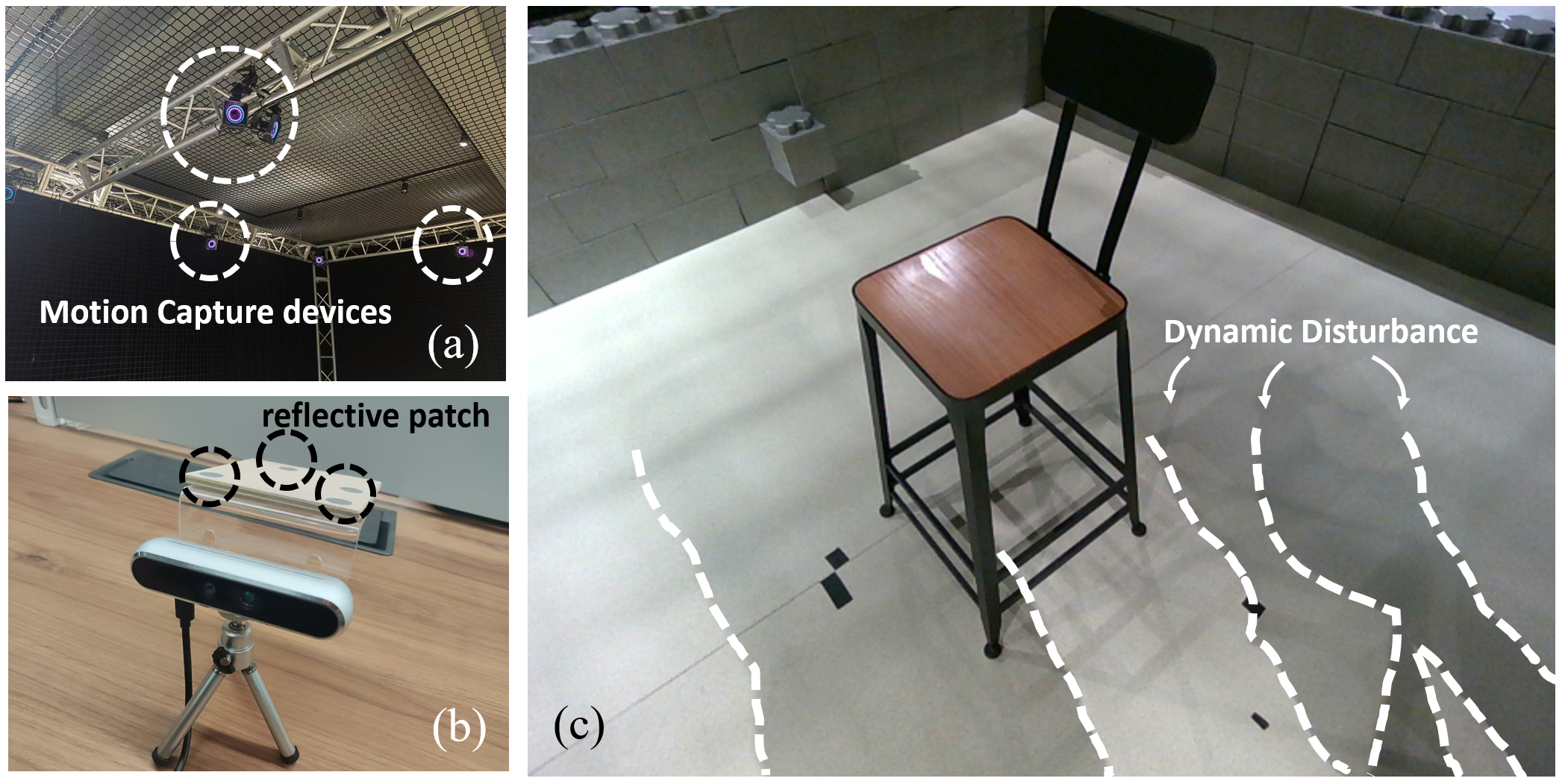}
\caption{Illustration of dataset collection. (a) The environment where the dataset is collected is equipped with a motion capture system. (b) The camera used to collect datasets. (c) The impression of the collected datasets, with images containing many dynamic human shadows.}
\label{dataset_collection}
\end{figure}

To further validate the generalization of our algorithm, we recorded our own dataset for comparison. Considering the impact of inaccurate depth measurements on pose estimation, we used the latest RealSense D455F depth camera to capture the RGB-D sequences. This camera provides an effective depth measurement range of up to six meters and has lower noise compared to the Kinect V1 used in the TUM datasets. Additionally, it features a global shutter and synchronized RGB and Depth channels. The data collection was conducted in an indoor environment equipped with a motion capture system. The camera and recording setup are shown in Fig. \ref{dataset_collection} (a) and Fig. \ref{dataset_collection} (b). 

\begin{figure}[h]
\centering
\includegraphics[width=0.9\linewidth]{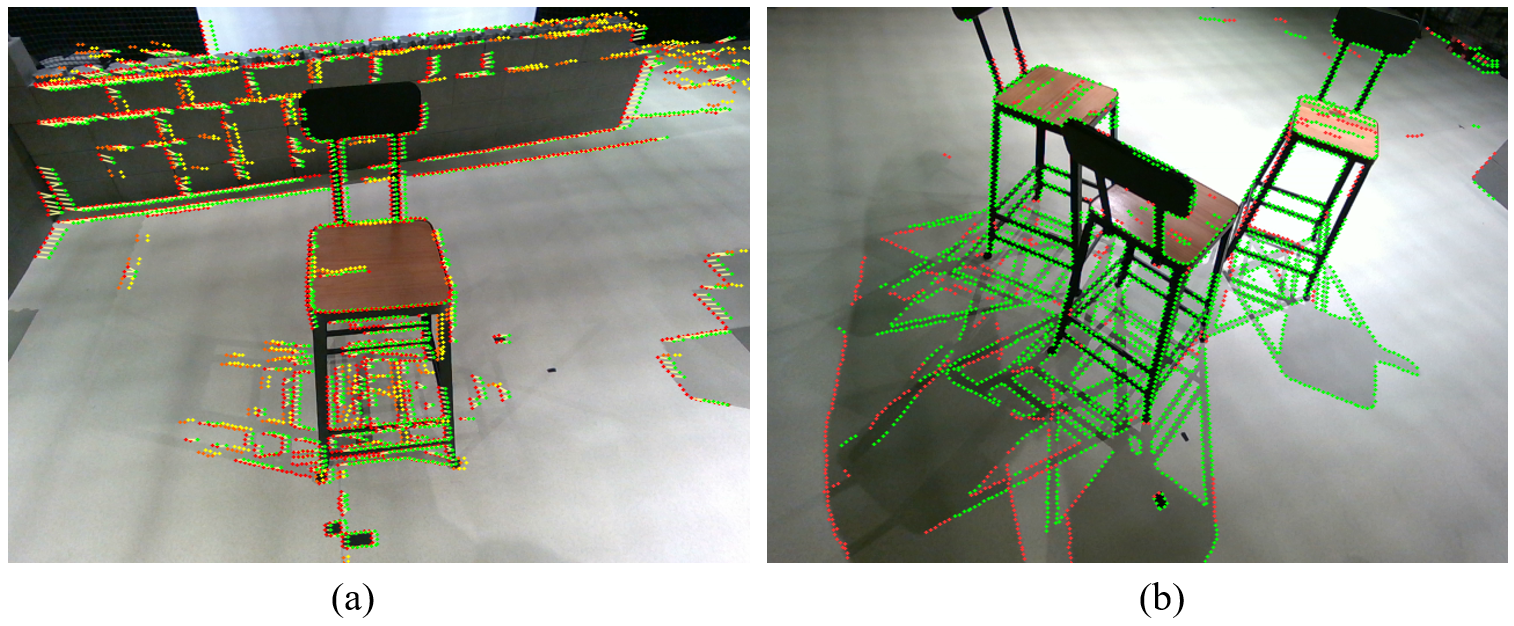}
\caption{Performance on the chair datasets. (a) Tracking process in \textit{single\_chair\_with\_loop}. (b) An illustration of edge authenticity distinction during the local mapping process in \textit{multi\_chair\_with\_loop}, with identified fake edges in red and real edges in green.}
\label{chair_run}
\end{figure}

\begin{figure}[h]
\centering
\includegraphics[width=0.9\linewidth]{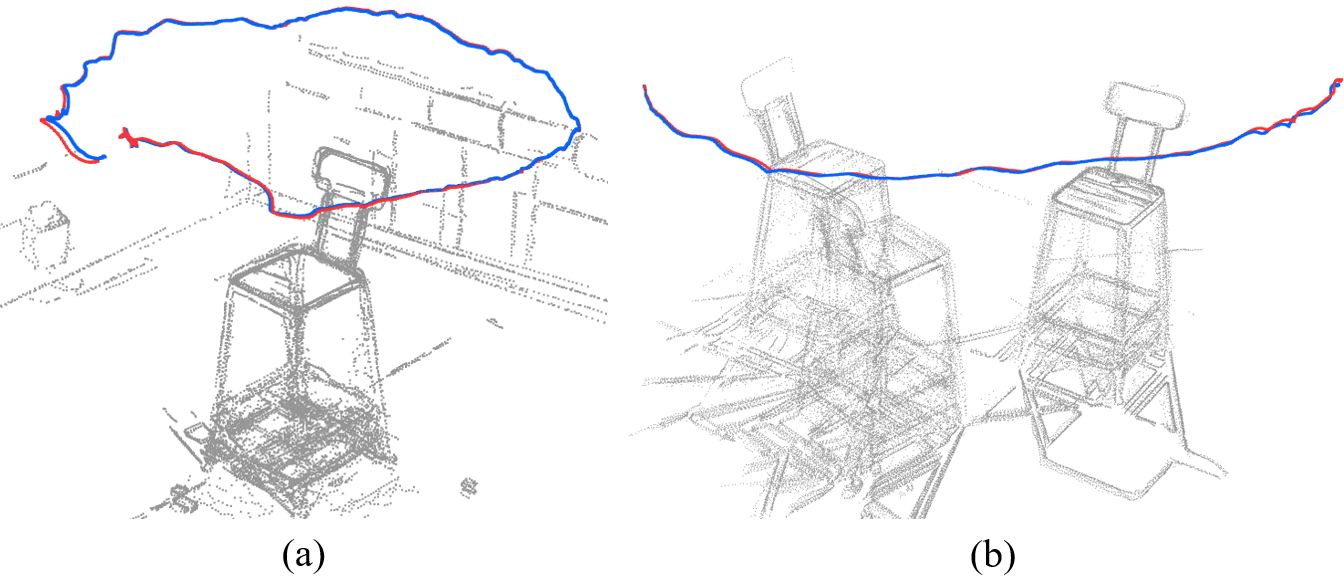}
\caption{Trajectory and semi-dense reconstruction results from Chair datasets, where (a) \textit{single\_chair\_with\_loop} and (b) \textit{multi\_chair\_no\_loop}. }
\label{chair_traj}
\end{figure}

We recorded sequences in scenes containing some chairs with significant dynamic human shadow disturbances as shown in Fig. \ref{dataset_collection} (c). Unlike the TUM RGB-D dataset, this dataset lacks artificially designed rich textures but only includes common structures and disturbances typical in real-world environments. We performed the same experiments on these sequences, with the algorithm's execution process shown in Fig. \ref{chair_run} and the resulting trajectories illustrated in Fig. \ref{chair_traj}. The trajectory errors are presented in Table \ref{ate_chair} and \ref{rpe_chair}.

\begin{table}[t]
\caption{Absolute Trajectory RMSE (cm) on Chair datasets\label{ate_chair}}
\centering
\setlength\tabcolsep{4pt}
\begin{tabular}{c|cccc|c}
\toprule
Method/Seq. & 1\_chair & 1\_chair & 3\_chairs & 3\_chairs & avg. \\
 & \_loop & \_no\_loop & \_loop & \_no\_loop \\
  \cmidrule{1-6}
  ORB-SLAM3-TR & 5.775 &  2.508 & 18.33 & 1.756 & 7.092\\
  ORB-SLAM3-LM\cite{orb-slam3} & 1.732 &  1.498 & 2.544 & 1.031 & 1.701 \\
 ManhattanSLAM\cite{mast3r-slam} & \second{1.264} &  \third{1.003} & \second{2.123} & \second{0.727} & \second{1.279}\\
 RE-SLAM\cite{re_slam} & 2.334 &  1.845 & 3.502 & 1.255 & 2.234\\
 DROID-SLAM\cite{droid-slam} & 2.302 &  1.539 & 2.586 & 1.333 & 1.940\\
 PHOTO-SLAM\cite{photo-slam} & 1.815 &  1.828 & \first{1.834} & \third{0.966} & \third{1.611}\\
 MAST3R-SLAM\textsuperscript{*}\cite{mast3r-slam} & 4.423 &  1.205 & 4.921 & 1.097 & 2.912\\
 \cmidrule{1-6}
 ROEVO-CT & 4.837 &  1.245 & 4.766 & 1.629 & 3.119\\
 ROEVO-TR & \third{1.722} &  \second{0.707} & 3.258 & 0.987 & 1.668\\
 ROEVO-TR+BA & \first{1.029} &  \first{0.655} & \third{2.323} & \first{0.476} & \first{1.121}\\
\bottomrule
\end{tabular}
\end{table}

\begin{table*}[!t]
\caption{Relative Trajectory RMSE on Chair Datasets (R: deg/s, t: cm/s)}
\label{rpe_chair}
\centering
\setlength\tabcolsep{3pt}
\begin{tabular}{c|cc|cc|cc|cc|cc}
\toprule
  & \multicolumn{2}{c}{1\_chair\_loop} & \multicolumn{2}{c}{1\_chair\_no\_loop} & \multicolumn{2}{c}{3\_chairs\_loop} & \multicolumn{2}{c}{3\_chairs\_no\_loop} & \multicolumn{2}{c}{average} \\
 
\cmidrule{1-11}
Seq. & RMSE (R) & RMSE (t)  & RMSE (R) & RMSE (t) & RMSE (R) & RMSE (t) & RMSE (R) & RMES (t) & RMSE (R) & RMSE (t)\\
\cmidrule{1-11}

ORB-SLAM3-TR\cite{orb-slam3} & 1.487 & 2.881 & 0.593 & 1.161 & 0.995 & 1.776 & 0.776 & 1.458 & 0.962 & 1.819\\
ORB-SLAM3-LM\cite{orb-slam3} & \first{0.511} & 0.751 & 0.461 & 0.695 & 0.682 & 1.103 & \second{0.344} & \second{0.686} & \second{0.499} & 0.809\\
 ManhattanSLAM\cite{manhattanslam} & \second{0.551} & 0.605 & 0.485 & 0.481 & 0.675& 1.154 & 0.745 & 0.971 & 0.614 & \third{0.803} \\
 RE-SLAM\cite{re_slam}  & 0.601 & 0.747 & 0.499 & 1.401 & 0.706 & 1.586 & 0.629 & 1.213 & 0.609 & 1.237 \\
 DROID-SLAM\cite{droid-slam} & \third{0.561} & \second{0.439} & \first{0.424} & \second{0.449} & 3.711 & 2.301 & 0.565 & 0.926 & 1.315 & 1.029\\
 PHOTO-SLAM\cite{photo-slam}  & 0.568 & 0.774 & \second{0.444} & 0.723 & \third{0.616} & \third{1.000} & 0.752 & 0.862 & 0.595 & 0.839\\
 MAST3R-SLAM\textsuperscript{*}\cite{mast3r-slam} & 0.613 & 7.010 & 0.566 & 0.633 & 0.776 & 2.841 &  0.587 & 1.383 & 0.635 & 2.967\\
 \cmidrule{1-11}
 ROEVO-CT & 0.689 & 0.707 & 0.533 & 0.674 & 0.815 & 1.048 & 0.627 & 0.994 & 0.666 & 0.856\\
 ROEVO-TR & 0.649 & \third{0.462} & 0.674 & \third{0.465} & \second{0.613} & \first{0.898} & \third{0.426} & \third{0.755} & \third{0.591} & \second{0.645} \\
 ROEVO-TR+BA & 0.577 & \first{0.344} & \third{0.465} & \first{0.383} & \first{0.597} & \second{0.968} & \first{0.328} & \first{0.641} & \first{0.492} & \first{0.584}\\
\bottomrule
\end{tabular}
\end{table*}

From Table \ref{ate_chair}, we can see that our method outperforms other methods on the chair dataset. Due to the sparse texture and dynamic disturbances such as human shadows, ORB-SLAM3 performs poorly on these sequences. In contrast, Manhattan-SLAM, which makes better use of the scene's structural features more effectively, achieves better results. We observed that in these datasets, photometric errors alone are not reliable for accurate pose estimation. However, our method, which effectively leverages both structure and texture, achieves optimal results.

Additionally, our method demonstrates higher robustness to dynamic disturbances in these sequences. As shown in Fig. \ref{chair_run} (b), by removing pseudo-edges during the association process and the local mapping process, our method effectively mitigates dynamic disturbances such as human shadows. This further enhances the trajectory quality as well as the reconstruction result. As seen in Fig. \ref{chair_traj}, the semi-dense reconstruction results are free from the influence of human shadows.

\subsubsection{Quantitative Sequence Summary}
Through the aforementioned evaluations, we comprehensively evaluate the performance of our method across four quantitative datasets and provide experimental results for multiple representative sequences from each benchmark. As shown in Table \ref{rank}, we evaluate the ATE rank over a total of 24 sequences across these four benchmarks. The table records the number of sequences in which each method ranks among the top three in terms of ATE.
\begin{table}[h]
\caption{Performance Ranking Comparison of ATE}
\label{rank}
\centering
\setlength\tabcolsep{6pt}
\begin{tabular}{c|ccc}
\toprule
Method/Seq Num. & Top 1 seqs. & Top 2 seqs. & Top 3 seqs. \\
  \cmidrule{1-4}
  ORB-SLAM3\cite{orb-slam3} & \third{3} & \second{9} & \second{13} \\
 ManhattanSLAM\cite{manhattanslam} & 1 & 6 & 8\\
 RE-SLAM\cite{re_slam} & 1  & 1& 5\\
 DROID-SLAM\cite{droid-slam} & 2 & 4 & 7\\
 PHOTO-SLAM\cite{photo-slam} & \second{4} & \second{9} & \second{13}\\
 MAST3R-SLAM\textsuperscript{*}\cite{mast3r-slam} & 0 & 1 & 1\\
 \cmidrule{1-4}
 ROEVO & \first{13} & \first{15} & \first{18}\\
\bottomrule
\end{tabular}
\end{table}

From Table \ref{rank}, we can see that our method outperforms all others in 13 out of 24 sequences in ATE, including 10 out of 20 sequences from public datasets, demonstrating strong cross-benchmark consistency.

\subsubsection{Outdoor Plant-Fence Dataset}
\begin{figure}[h]
\centering
\includegraphics[width=3 in]{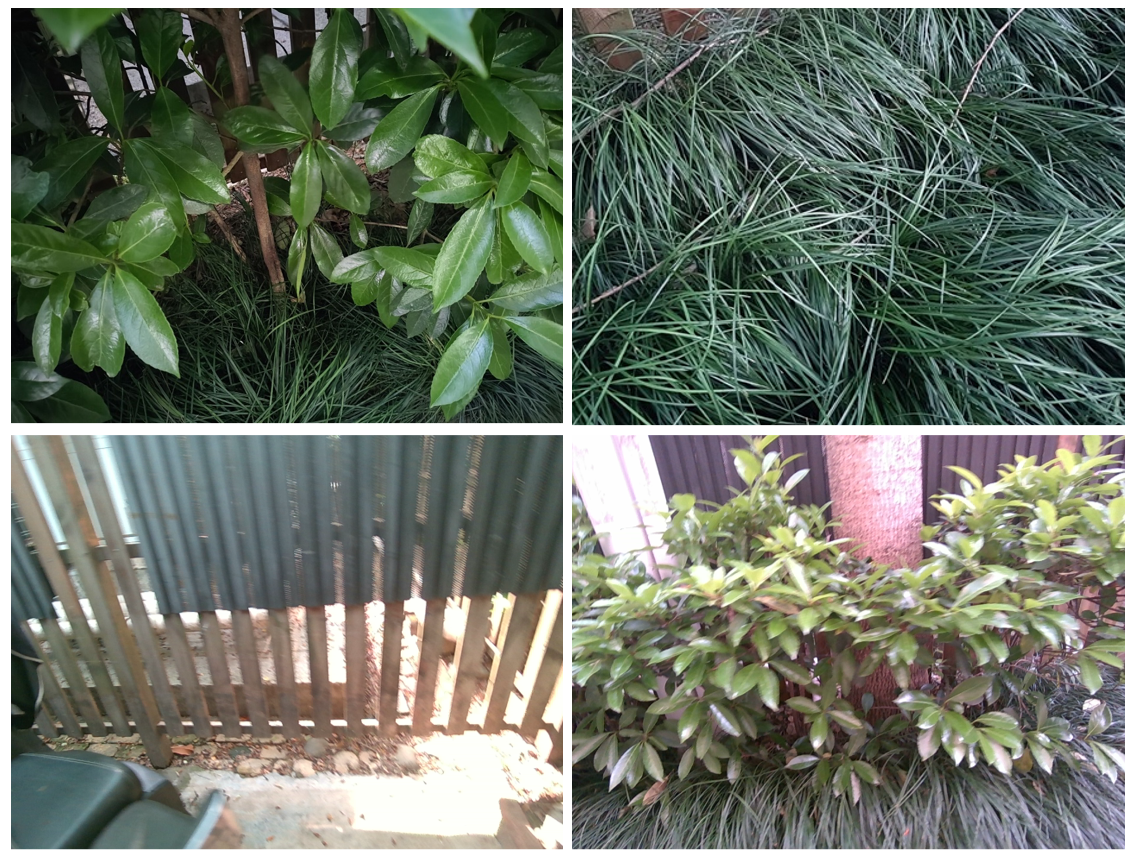}
\caption{Illustrations of the environment of the Plant-Fence sequence. The sequence contains a lot of frames with cluttered shrubs, dense weeds, and neat fence structures. The leaves exhibit reflections and extensive occlusion caused by viewpoint variations, while the fences exhibit a high degree of structural similarity. }
\label{outdoor_env}
\end{figure}

\begin{figure*}[h]
\centering
\includegraphics[width=0.85\linewidth]{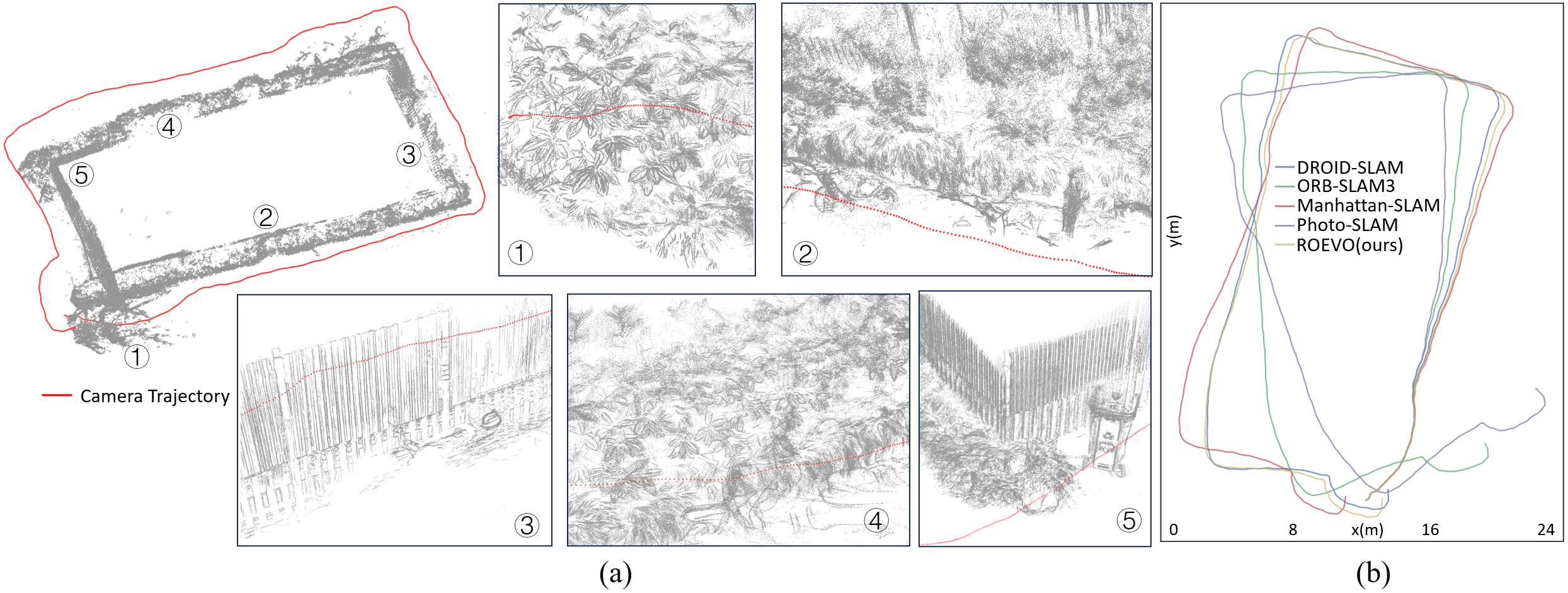}
\caption{Trajectories and semi-dense reconstruction results on the Plant-Fence dataset. (a) The reconstruction results and detailed examples in some typical areas using the proposed method. (b) The estimated trajectories by multiple methods.}
\label{outdoor_rec}
\end{figure*}

To further validate the generalization capability of our method, we also captured a large-scale outdoor sequence by scanning the surroundings of a large green belt, where the environment is illustrated in Fig. \ref{outdoor_env}. This scene is more challenging than typical indoor data for at least three reasons. First, the environment is an open outdoor scene covering approximately 500 $m^2$, with a trajectory length of approximately 150$m$ and highly variable illumination conditions. Second, the sequence contains dense shrubs and weeds, where leaves and grass frequently cause severe occlusion and reflections under camera viewpoint changes. Third, the abundance of near-identical fence structures in the scene significantly increases the difficulty of data association. The scanning was performed using a RealSense D455F camera, with strict timestamp synchronization between aligned depth maps and RGB images.
Due to the absence of ground-truth poses, we evaluated trajectory accuracy using loop closure error: the distance discrepancy between the start and end points when the camera returns to its initial position after scanning. The estimated trajectories of all algorithms are visualized in Fig. \ref{outdoor_rec} (b), with quantitative loop closure errors listed in Table \ref{loop_err}. We excluded mast3r-SLAM \cite{mast3r-slam} from quantitative evaluation on this sequence because it cannot provide metrically scaled trajectories without scale calibration. In such cluttered environments, methods that heavily rely on optimization across multiple frames like ORB-SLAM3 \cite{orb-slam3} and Photo-SLAM \cite{photo-slam} exhibit significant drift due to severe occlusion and illumination variations between frames that disrupted robust data association. Other edge-based methods like RE-SLAM \cite{re_slam} failed completely, this is because distance field(DF)-based methods often fail to converge in such environments with extreme edge clutter due to their inability to establish reliable data associations and effectively organize features. 
In contrast, our method that leverages organized edges for tracking and local mapping demonstrated superior advantages. Fig. \ref{outdoor_rec} (a) presents close-up views of the trajectory and reconstructed scene generated by our method. Notably, our approach successfully reconstructs well-aligned structures of cluttered leaves, weeds, and fences. This capability stems from the utilization of organized edges, which adapts to diverse edge shapes and leverages them for high-precision feature association and localization.

\begin{figure}[h]
\centering
\includegraphics[width=0.9\linewidth]{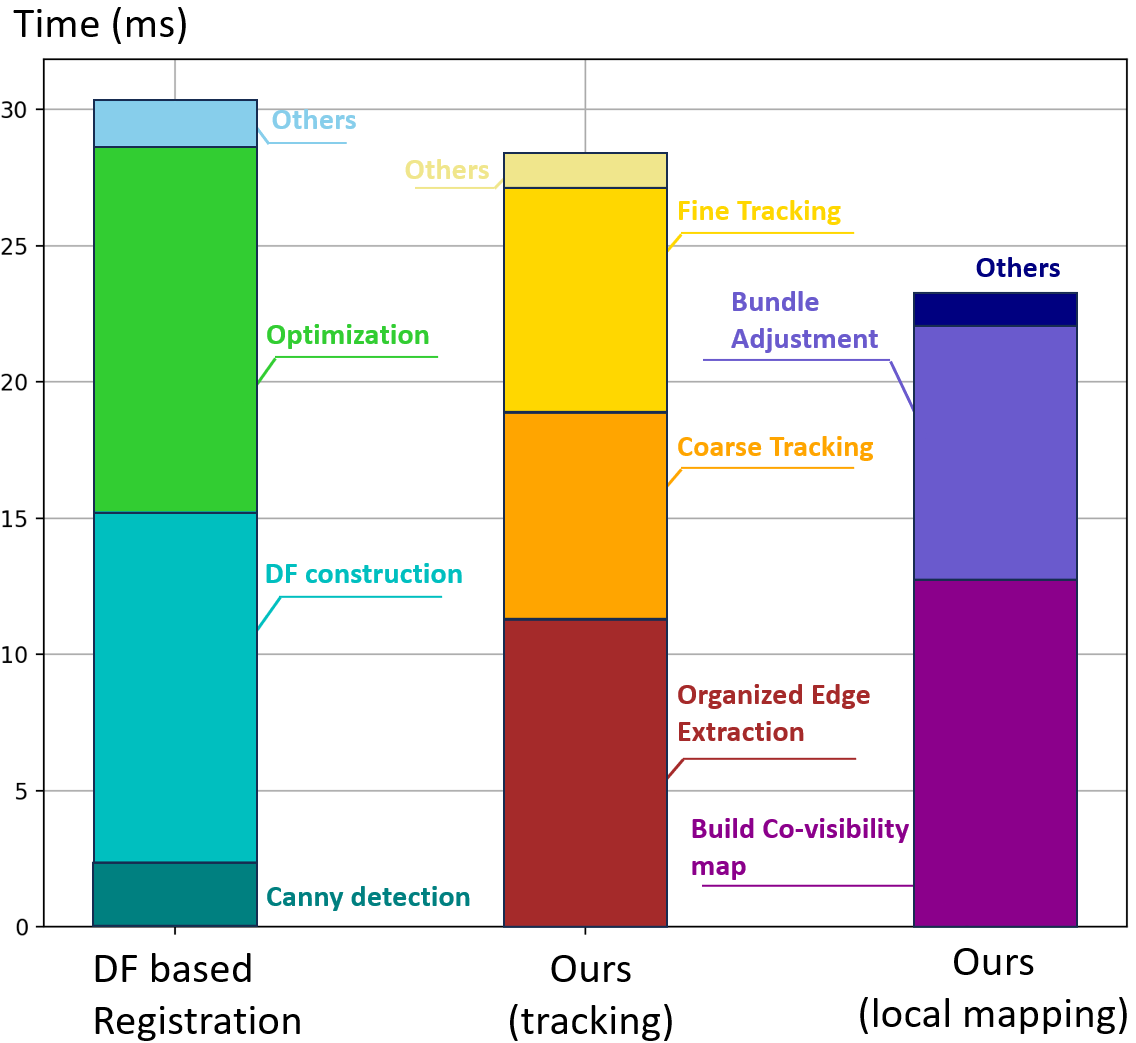}
\caption{Bar chart of algorithm runtime statistics, where each block corresponds to the time taken by a specific operation. }
\label{time}
\end{figure}

\begin{figure*}[h]
\centering
\includegraphics[width=0.85\linewidth]{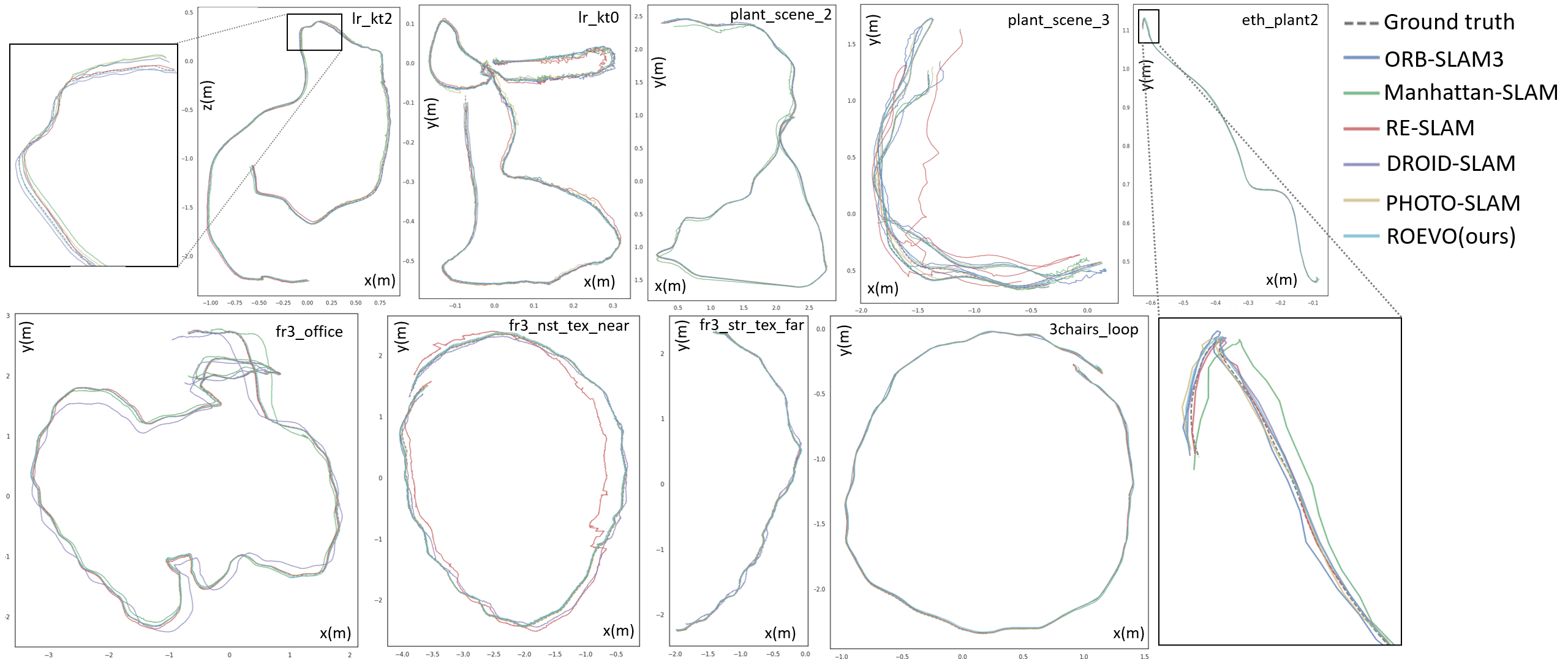}
\caption{The trajectories from various sequences, including those from the ICL-NUIM, ETH-3D, TUM-RGBD, and Chair datasets. The ground truth is represented by \textcolor{gray}{gray} dashed lines, while the proposed method's trajectories are depicted in \textcolor{cyan}{cyan}. All trajectories have been aligned with the ground truth for fair comparison. Our method consistently produces trajectories that closely match the ground truth across all sequences. }
\label{traj_total}
\end{figure*}

Through the aforementioned series of comparative experiments on benchmarks, it can be observed that our method demonstrates consistent performance across multiple benchmarks and achieves or surpasses state-of-the-art geometry-based and learning-based approaches on a number of sequences. Qualitative visualizations of the estimated trajectories for several selected sequences from these benchmarks are presented in Fig. \ref{traj_total}. As illustrated in the figure, compared to other methods that exhibit less stable cross-benchmark performance, such as DROID-SLAM \cite{droid-slam}, Manhattan-SLAM \cite{manhattanslam}, Re-SLAM \cite{re_slam}, or Photo-SLAM \cite{photo-slam}, our approach consistently produces trajectories that closely align with the ground truth while maintaining structural fidelity, thereby validating the robustness and consistency of our method.

\subsection{Ablation Studies}
In this subsection, we evaluate the accuracy improvements in pose estimation achieved by utilizing organized edges in coarse tracking. Specifically, we employ organized edges to compute photometric errors for direct tracking, along with a rotation-invariant photometric error defined based on edge gradients, as shown in Eq. \ref{photometric}. We compared four methods for direct tracking across multiple sequences: (1) using keypoints selection schemes proposed by DSO \cite{dso}, (2) using Canny edges \cite{canny} without any extra processing, (3) using organized edges without rotational-invariant photometric error, and (4) using organized edges with rotational-invariant photometric error. The Relative Pose Error (RPE) was used to evaluate their ability to estimate relative pose transformations. The results are presented in Table \ref{ablation}.
From the experimental results, it can be observed that conventional Canny edges introduce significant noise into the optimization process due to the indistinguishable edge pixels. In contrast, organized edges include complete and salient structural features in the environment yield superior pose estimation accuracy. Furthermore, by incorporating the rotation-invariant photometric error, the system achieves additional improvement in rotational estimation capability.

\begin{table*}[h]
\caption{Relative Trajectory RMSE (R: deg/s, t: cm/s) with Different Coarse Tracking Methods}
\label{ablation}
\centering
\setlength\tabcolsep{3pt}
\begin{tabular}{c|cc|cc|cc|cc}
\toprule
  Method & \multicolumn{2}{c}{features\_from\_DSO\cite{dso}} & \multicolumn{2}{c}{Raw\_Canny\cite{canny}} & \multicolumn{2}{c}{Organized\_Edges} & \multicolumn{2}{c}{Organized\_Edges+Er} \\
 
\cmidrule{1-9}
Metric & RMSE (R) & RMSE (t)  & RMSE (R) & RMSE (t) & RMSE (R) & RMSE (t) & RMSE (R) & RMES (t) \\
\cmidrule{1-9}

 &\second{0.538} &  \third{1.207} & 0.585 & 1.352 & \third{0.574} & \second{1.168} & \first{0.501} & \first{1.137} \\
\bottomrule
\end{tabular}
\end{table*}

\begin{table*}[t]
\caption{Loop-Closure Error (m) On Plant-Fence Sequence \label{loop_err}}
\centering
\setlength\tabcolsep{6pt}
\begin{tabular}{c|c|c|c|c|c}
\toprule
 ORB-SLAM3\cite{orb-slam3} & ManhattanSLAM\cite{manhattanslam} & RE-SLAM\cite{re_slam} & DROID-SLAM\cite{droid-slam} & PHOTO-SLAM\cite{photo-slam} & ROEVO \\
  \cmidrule{1-6}
   7.771 & \second{1.263} & $\times$ & \third{1.621} & 11.16 & \first{0.94} \\
\bottomrule
\end{tabular}
\end{table*}

\subsection{Repeatability Studies}

\begin{figure}[h]
\centering
\includegraphics[width=0.95\linewidth]{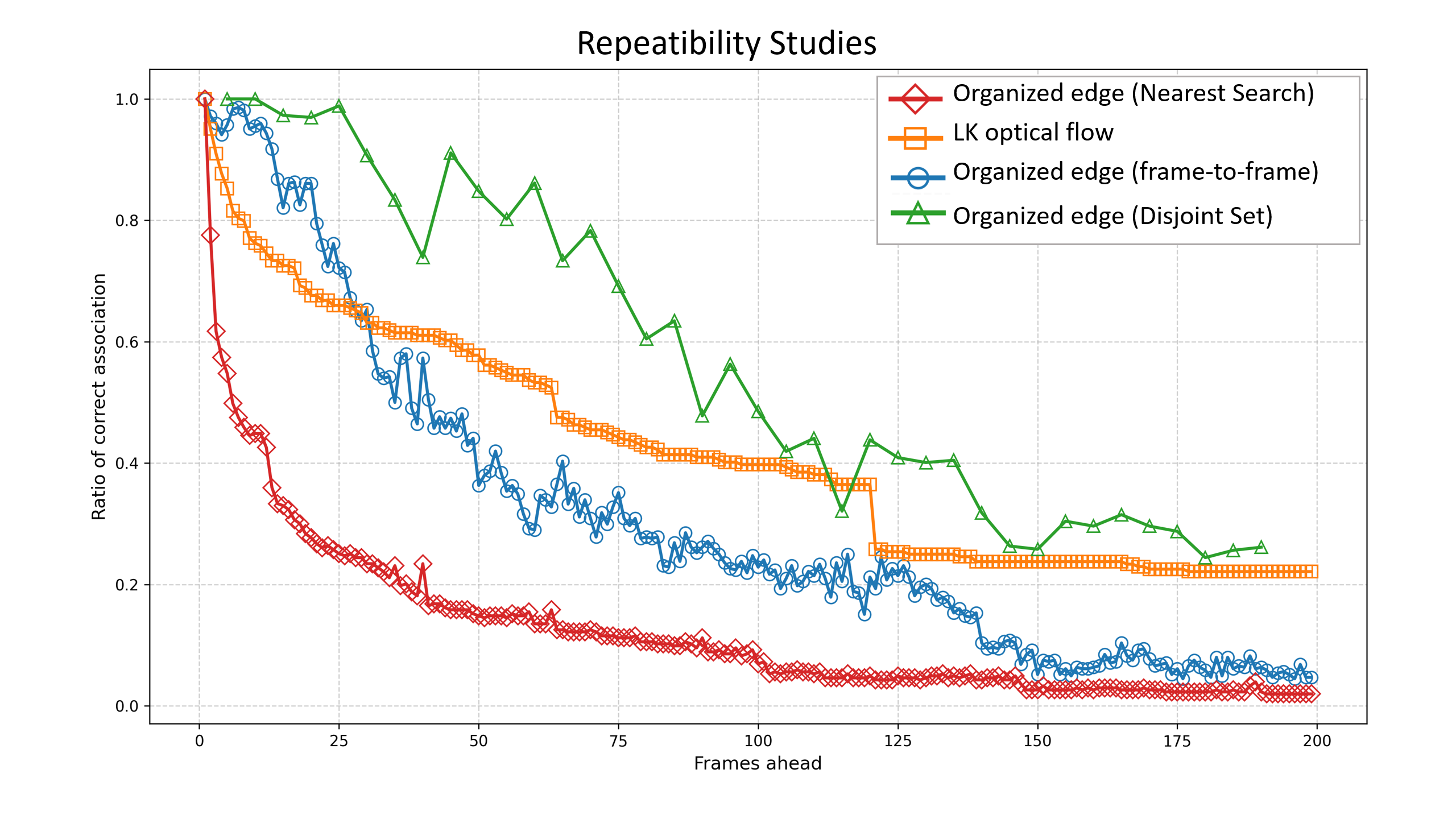}
\caption{Cross-frame association result using different methods, where results of organized edges with nearest neighbor marked in \textcolor{red}{red}, results of LK optical flow marked in \textcolor{orange}{orange}, frame-to-frame association with edge normal marked in \textcolor{blue}{blue}, and results of disjoint-set-based association in a co-visibility graph marked in \textcolor{green}{green}. }
\label{repeat}
\end{figure}

Since the organized edge is a class of associable features, we conduct experiments to investigate its repeatability under viewpoint variations. As organized edges are associated across continuous video streams, we selected three sequences from the TUM \cite{tum} dataset (\textit{fr1\_desk, fr2\_desk}, and \textit{fr3\_office}) for evaluation. The organized edges extracted from the first frame were established as the reference, and then we subsequently computed the proportion of correctly associated features in subsequent frames relative to this reference frame. As the proposed method discusses two distinct association methodologies: the first involves frame-to-frame association during the tracking process, as detailed in Section \ref{fine}, while the second employs a disjoint set-based co-visibility graph for multi-frame association, as presented in Section \ref{section_co}. We systematically evaluated both approaches. For comparative analysis, we benchmark the performance against two alternative methods: (1) Feature association through Shi-Tomasi corner points using Lucas-Kanade optical flow \cite{lkoptic}, and (2) Associating organized edges frame-by-frame utilizing original nearest-neighbor matching. The experimental results are illustrated in Fig. \ref{repeat}.
As illustrated in the figure, the frame-to-frame association demonstrates fine performance under limited viewpoint variations, outperforming the LK optical flow method in such scenarios. However, its effectiveness diminishes as the viewpoint changes become more pronounced, eventually falling below that of the LK optical flow approach. This observation suggests its particular suitability for tracking phases with constrained viewpoint variations.

Upon incorporating the disjoint-set-based association, the organized edges exhibit significantly improved association performance, surpassing the LK optical flow method. This enhancement can be attributed to the disjoint-set mechanism, which consolidates multi-frame association information, thereby yielding more robust and coherent matching results. Furthermore, the association method specifically designed for organized edges consistently outperforms the raw nearest-neighbor approach, further validating the efficacy of the proposed methodology.

\subsection{Efficiency Analysis}

Real-time performance is a critical aspect of any VO system. To evaluate the computational efficiency of our method, we conducted tests on a laptop equipped with an Intel Core i7-12700H CPU. For a fair comparison, we also implemented a distance field (DF)-based registration scheme \cite{8ssedt} on the same computer with identical configurations. The test results are presented in Fig. \ref{time}. Since local mapping and tracking operate on separate threads, we evaluated their performance individually. The local mapping thread, which utilizes keyframes and maintains a sliding window during bundle adjustment, does not require processing every RGB-D frame. Therefore, we calculated the average time cost per frame for local mapping.

Fig. \ref{time} illustrates that our method outperforms distance field-based approaches in terms of runtime overhead, maintaining a frequency of approximately 35 Hz during practical use. While the optimization efficiency of our method is comparable to that of standard registration approaches, our feature processing is faster than the creation of distance fields. Additionally, the average efficiency of our local mapping thread surpasses that of the tracking thread, further demonstrating that the two threads can operate in parallel during real-time execution.

\section{Conclusions}

In this paper, we propose ROEVO, a VO system that utilizes organized edge features to effectively leverage the structural and textural information of image edges. We have developed a complete VO pipeline by exclusively utilizing organized edges, which integrates intensity, texture, and structural information of image edges comprehensively into data association, tracking, local mapping, and bundle adjustment. Our method demonstrates significant accuracy and robustness in indoor environments. Through practical experiments, we show that our approach exhibits generalizable patterns in selecting configurations and incurs reasonable computational overhead, making it a versatile solution for pose estimation based on edge features. While our current method remains susceptible to image motion blur and exhibits reduced accuracy in scenarios with abundant pseudo-edges, we plan to address these corner cases in future work through potential integration of IMU data or by incorporating surface/planar features. Our method can also be extended to monocular or stereo settings in the future. With further development, it has the potential to evolve into a more complete SLAM system capable of handling large viewpoint variations.

\section{Appendix}

\subsection{Jacobian Derivation}\label{app_j}
In this section, we present the analytical form of the Jacobian matrix for the 3D-2D registration problem. We adopt the notation used in Section \ref{fine}, that is, we compute $\partial p'/\partial \xi$ in Equation (\ref{fine_jacobian}), where $\xi$ represents the 6-DOF pose and $p'$ is the reprojected pixel point.

We use the commonly adopted method in optimization problems to represent translation and rotation \cite{lieg}, which involves utilizing the exponential map $\exp(\cdot)$ to perturb the rotation. This allows us to incrementally optimize the rotation during the optimization process, \textit{i.e.}, $\mathbf{R}_{i+1} = \mathbf{R}_{i}\exp(\delta\boldsymbol{\phi}^{\land})$, where $\delta\boldsymbol{\phi}$ is the 3-DOF rotational increment, and $()^{\land}$ denotes the operation that forms a $3\times 3$ skew-symmetric matrix from a 3D vector. $\boldsymbol{\phi}$  map itself into a $3\times3$ rotation through Rodrigues' formula:
\begin{equation}
    \label{Rodrigues}
    \exp(\boldsymbol{\phi}^{\land}) = \mathbf{I} + \frac{1 -\cos\Vert\boldsymbol{\phi}\Vert}{\Vert\boldsymbol{\phi}\Vert^2}(\boldsymbol{\phi}^{\land})^2+\frac{\sin\Vert\boldsymbol{\phi}\Vert}{\Vert\boldsymbol{\phi}\Vert}\boldsymbol{\phi}^{\wedge}.
\end{equation}

Additionally, according to the derivations in \cite{lieg,vio}, for a 3D rotational problem $\mathbf{p} = \mathbf{R}\mathbf{p}_0$, we can compute the Jacobian of a rotation perturbation $\delta \boldsymbol{\phi}$ by $(\partial \exp(\boldsymbol{\phi})\mathbf{p}_0)/(\partial \delta\boldsymbol{\phi}
) = -(\mathbf{R}\mathbf{p}_0)^{\land}$, which allows us to efficiently compute the incremental rotation during the optimization process. Similarly, for a complete pose transformation with both translation and rotation like $\mathbf{p} = \mathbf{R}\mathbf{p}_0 + \mathbf{t}$, we can define a 6-DOF pose transformation as $\boldsymbol{\xi} = [\mathbf{t}, \boldsymbol{\phi}]$, and the corresponding Jacobian with respect to the pose increment is then given by:
\begin{equation}
    \label{dp_ddxi}
    (\partial \exp(\boldsymbol{\xi})\mathbf{p}_0)/(\partial \delta\boldsymbol{\xi}) = [\mathbf{I}_{3\times3}, -(\mathbf{R}\mathbf{p}_0)^{\land}]
\end{equation}
Then, we can apply the chain rule to obtain $\partial p'/\partial \delta\boldsymbol{\xi}$ as:
\begin{equation}
    \label{chain}
    \frac{\partial p'}{\partial \delta\boldsymbol{\xi}} = \frac{\pi(\exp(\boldsymbol{\xi}^{\land})\mathbf{p}_0)}{\partial \delta\boldsymbol{\xi}}=\frac{\partial\pi(\mathbf{p})}{\partial \mathbf{p}}\frac{\partial \exp(\xi^{\land})\mathbf{p}_0}{\partial \delta\boldsymbol{\xi}}
\end{equation}
where $\frac{\partial\pi(\mathbf{p})}{\partial \mathbf{p}}$ is the Jacobian matrix of the camera projection, given by:
\begin{equation}
    \label{ja_proj}
   \frac{\partial\pi(\mathbf{p})}{\partial \mathbf{p}} = 
\begin{bmatrix}
f_x/\mathbf{p}_z & 0 & -f_x\mathbf{p}_x/(\mathbf{p}_z)^2 \\
0 & f_y/\mathbf{p}_z & -f_y\mathbf{p}_y/(\mathbf{p}_z)^2
\end{bmatrix}
\end{equation}
where $\mathbf{p} = \exp(\xi^{\land})\mathbf{p}_0$. Thus, the final expression for $\partial p'/\partial \delta\boldsymbol{\xi}$ can be written as:
\begin{equation}
    \label{j_3d_2d_t}
(\frac{\partial p'}{\partial \delta\boldsymbol{\xi}})_{(:,1:3)} = 
\begin{bmatrix}
\frac{f_x}{\mathbf{p}_z} & 0 & -f_x\frac{\mathbf{p}_x}{(\mathbf{p}_z)^2}  \\
0 & \frac{f_y}{\mathbf{p}_z} & -f_y\frac{\mathbf{p}_y}{(\mathbf{p}_z)^2} 
\end{bmatrix}
\end{equation}

\begin{equation}
    \label{j_3d_2d_r}
(\frac{\partial p'}{\partial \delta\boldsymbol{\xi}})_{(:,4:6)} = 
\begin{bmatrix}
 -f_x\frac{\mathbf{p}_x\mathbf{p}_y}{{\mathbf{p}_z}^2} & f_x + f_x\frac{{\mathbf{p}_x}^2}{{\mathbf{p}_z}^2} & -f_x\frac{\mathbf{p}_y}{\mathbf{p}_z}\\
 -f_y - f_y\frac{{\mathbf{p}_y}^2}{{\mathbf{p}_z}^2}  & f_y\frac{\mathbf{p}_x\mathbf{p}_y}{{\mathbf{p}_z}^2} & -f_y\frac{\mathbf{p}_x}{\mathbf{p}_z}
\end{bmatrix}
\end{equation}

\subsection{Algorithm of Constructing Covisibility Graph}\label{app_alg}

\begin{algorithm}[H]
\caption{Initialize a Co-visibility Graph }\label{algorithm_init}
\begin{algorithmic}
\STATE 
\STATE \textbf{Input} Key frame $\boldsymbol{F}_{r}$ that contain $ = \{ ^{r}\mathcal{E}_1, ^{r}\mathcal{E}_2, ... , ^{r}\mathcal{E}_N\}$
\STATE \textbf{Output} Disjoint set $\mathcal{D}$, Keyframe list $\mathcal{K}$
\STATE \hspace{0.5cm} Disjoint set $\mathcal{D} \gets \varnothing$
\STATE \hspace{0.5cm} Keyframe stack $\mathcal{K} \gets \varnothing$
\STATE \hspace{0.5cm} \textbf{for} $^{ref}\mathcal{E}_1 \in \mathcal{F}_{ref}$ 
, $i = 0, 1, ..., N$, \textbf{do:}
\STATE \hspace{1.0cm} \textit{tuple} $\boldsymbol{e}$ $\gets$ $(\boldsymbol{F}_{r}.\mathbf{ID}, i)$
\STATE \hspace{1.0cm} $\mathcal{D}$.\textbf{insert}($\boldsymbol{e}$)
\STATE \hspace{1.0cm} $\mathcal{D}$.\textbf{locate}($\boldsymbol{e}$) $\gets$ $\boldsymbol{e}$
\STATE \hspace{0.5cm} $\mathcal{K}$.\textbf{push}($\boldsymbol{F}_{r}$)
\STATE \hspace{0.5cm} \textbf{return} $\mathcal{D}$, $\mathcal{K}$ 
\end{algorithmic}
\label{alg1}
\end{algorithm}

{\small
\bibliographystyle{IEEEtran}
\bibliography{refsota, ref, refbenchmark}
}

\end{document}